\documentclass[10pt,twocolumn,letterpaper]{article}

\usepackage{cvpr}              
\usepackage[dvipsnames]{xcolor}

\usepackage{graphicx}
\usepackage{amsmath}
\usepackage{amssymb}
\usepackage{booktabs}

\usepackage{tikz}
\usepackage{comment}
\usepackage{color}
\usepackage{nccmath}
\usepackage{soul}

\usepackage{algorithm}
\usepackage{algpseudocode}

\usepackage{enumitem}
\definecolor{myyellow}{RGB}{176,149,0}
\definecolor{myred}{RGB}{178,0,0}
\definecolor{myblue}{RGB}{0, 29, 188}
\definecolor{mygreen}{RGB}{45, 118, 0}

\newcommand{\alan}[1]{\textcolor{red}{}}
\newcommand{\adam}[1]{\textbf{\textcolor{red}{}}}
\newcommand{\yaoyao}[1]{\textcolor[rgb]{0.12, 0.1, 0.86}{}}
\newcommand{\prakhar}[1]{\textcolor{purple}{}}

\newcommand{\elasticgauss}{
\begin{table*}
\setlength{\tabcolsep}{4pt}
\centering
\caption{Imagenet-C Corruptions on Pascal3D+ dataset - Classification Results (Vgg16)}
\label{table:elasticgauss}
\begin{tabular}{lcccc|cccc|cccc}
\hline\noalign{\smallskip}
\textbf{Model} & \multicolumn{4}{c}{\textbf{Elastic Transform}} & \multicolumn{4}{c}{\textbf{Gaussian Blur}} & \multicolumn{4}{c}{\textbf{Snow}} \\
\textit{Occlusion}$\rightarrow$  &\small {\textbf{0\%}} & \small {\textbf{20-40\%}} & \small {\textbf{40-60\%}} &  \small {\textbf{60-80\%}} & \small {\textbf{0\%}} & \small {\textbf{20-40\%}} & \small {\textbf{40-60\%}} &  \small {\textbf{60-80\%}} & \small {\textbf{0\%}} & \small {\textbf{20-40\%}} & \small {\textbf{40-60\%}} &  \small {\textbf{60-80\%}}  \\
\noalign{\smallskip}
\hline
\noalign{\smallskip}                                                                        
\textbf{RPL~\cite{rpl}} & .830           & .597          & .461          & .371          & .855            & .541         & .403          & .320          & .842              & .592           & .435            & .408\\
\textbf{BNA~\cite{bna}} & .793            & .601            & .498            & .400            & .833              & .618           & .484            & .300           & .767              & .627           & .542            & .454\\  
\hdashline
\textbf{CompNet~\cite{compnet2020journal}}& .268          & .183          & .157          & .146          & .732            & .395         & .296          & .241         & .529 & .348           & .258            & .210\\    
\textbf{UGT (Ours)}    & \textbf{.872} & \textbf{.712} & \textbf{.712} & \textbf{.494} & \textbf{.909}   & \textbf{.720} & \textbf{.613} & \textbf{.509}& \textbf{.890} & \textbf{.742}& \textbf{.634}&\textbf{ .523}\\
\hline
 & \multicolumn{4}{c}{\textbf{Motion Blur}} & \multicolumn{4}{c}{\textbf{Contrast}} & \multicolumn{4}{c}{\textbf{Frost}} \\
\hline
\textbf{RPL~\cite{rpl}} & .862 &.629 &.481 &.373        & .901 &.610 &.433 &.321 & .850 &.670 &.511& .402\\       
\textbf{BNA~\cite{bna}}         & .844 & .623             & .481         & .355          & .899 & .601&.401&.315&.845&.654&.501&.399\\
\hdashline
\textbf{CompNet~\cite{compnet2020journal}}     & .639 & .362 & .287 & .241 & .760 &.472 &.374 &.312 & .740 & .481 &.360 &.301\\ 
\textbf{UGT (Ours)} & \textbf{.891} &\textbf{.763} &\textbf{.673} &\textbf{.567} & \textbf{.923} &\textbf{.701} &\textbf{.534} &\textbf{.412} & \textbf{.911} &\textbf{.782} &\textbf{.672} &\textbf{.561}\\
\hline
\end{tabular}
\end{table*}
}

\newcommand{\udaparttable}{
\begin{tiny}
\begin{table}[h]
\centering
\caption{Synthetic (UDAParts)~\cite{udapart22} to Real (Pascal3D+)~\cite{pascal3d14} dataset - Classification Results on Resnet50}
\label{table:udaparts}
\begin{tabular}{llcccc}
\hline\noalign{\smallskip}
\textbf{Model}       & \small {\textbf{0\%}} & \small {\textbf{20-40\%}} & \small {\textbf{40-60\%}} &  \small {\textbf{60-80\%}}\\
\noalign{\smallskip}
\hline
\noalign{\smallskip}
\textbf{RPL~\cite{rpl}}         & .822            & .432         & .370         & .335\\       
\textbf{BNA~\cite{bna}}         & .950            & .684         & .484         & .356\\
\hdashline
\textbf{CompNet~\cite{compnet2020journal}}      & .940         & .650         & .475         & .347\\ 
\textbf{UGT (Ours)}            & \textbf{.992}    & \textbf{.957} & \textbf{.861} & \textbf{.753} \\
\hline
\vspace{-9mm}
\end{tabular}
\end{table}
\end{tiny}
}

\newcommand{\totalandudatable}{
\begin{tiny}
\begin{table*}

\setlength{\tabcolsep}{4pt}
\centering
\caption{OOD-CV Nuisances Top-1 Classification Results. Occlusion levels greater than $0\%$ represent Occluded-OOD-CV dataset.}
\label{table:robintotaluda}
\begin{threeparttable}
\begin{tabular}{lcccc|cccc|cccc}
\hline\noalign{\smallskip}
\textbf{Method} & \multicolumn{4}{c}{\textbf{Combined}} & \multicolumn{4}{c}{\textbf{Context}} & \multicolumn{4}{c}{\textbf{Weather}} \\
\textit{Occlusion}$\rightarrow$  &\small {\textbf{0\%}} & \small {\textbf{20-40\%}} & \small {\textbf{40-60\%}} &  \small {\textbf{60-80\%}} & \small {\textbf{0\%}} & \small {\textbf{20-40\%}} & \small {\textbf{40-60\%}} &  \small {\textbf{60-80\%}} & \small {\textbf{0\%}} & \small {\textbf{20-40\%}} & \small {\textbf{40-60\%}} &  \small {\textbf{60-80\%}}  \\
\noalign{\smallskip}
\hline
\noalign{\smallskip}
\textbf{CDAN~\cite{cdan18}}\tnote{**}   & .760      & .531            & .420            & .380            & .710           & .541           & .436             & .397            & .745           & .476           & .335             & .299\\
\textbf{BSP~\cite{bsp19}}\tnote{**}     & .753     & .506            & .401            & .351            & .610           & .511           & .419             & .385            & .730           & .391           & .266             & .254\\
\textbf{MDD~\cite{mdd19}}\tnote{**}     & .780     & .551            & 469            & .410            & .761           & .531           & .436             & .410            & .802           & .439           & .306             & .271\\
\textbf{MCD~\cite{mmcd18}}\tnote{**}    & .772     & .556            & .461            & .403            & .798           & .523           & .426             & .374            & .810           & .447           & .336             & .286\\
\textbf{MCC~\cite{mcc19}}\tnote{**}     & .785     & .582            & .492            & .434            & .730     & .577            & .454            & .420          & .767           & .503           & .376             & .362\\  
\textbf{FixBi~\cite{na2021fixbi}}\tnote{**}       &  .821    & .534     &  .478         & .399          &    .802 & .542 & .445 & .409       &   .755         &  .489          & .358             & .335\\
\textbf{MIC~\cite{hoyer2023mic}}\tnote{**}        & .837    & .540     &  .376         &  .262         &    .755 & .602 & .532 & .499       &  .817          & \textbf{.612}           & .496              & .427 \\
\textbf{ToAlign~\cite{wei2021toalign}}\tnote{**}  & .761     & .507     & .411          & .346          &    .712 & .501 & .393 & .382       &   .720         & .381           & .252             &.213 \\
\textbf{CST~\cite{liu2021cyclecst}}\tnote{**}     & .840     & .579     &  .539         & .477          &    .687 & .491 & .452 & .411       &   .813         &  .558          &     .397         & .356\\
\textbf{DUA~\cite{mirza2022normdua}}\tnote{**}    &  .699    & .523     &   .480        & .403          &    .667 & .471 & .434 & .401       &   .701         & .465           & .391             &.210 \\
\textbf{DINE~\cite{liang2022dine}}\tnote{**}      & .835     & .600     &  .493         &  .443         &    .867 & .515 & .418 & .397       &   .798         &  .423          &  .290            & .261\\
\hdashline
\textbf{RPL~\cite{rpl}}                      & .664               & .430          & .346          & .300          & .675           & .457          & .368           & .315          & .642           & .247           & .138             & .122\\       
\textbf{BNA~\cite{bna}}                      & .653                & .426          & .343          & .298          & .580           & .397           & .342           & .278            & .635           & .295           & .179             & .171\\
\hdashline
\textbf{CompNet~\cite{compnet2020journal}}   & .720               & .506          & .462          & .415          & .790           & .517           & .454             & .369            & .683           & .434           & .398             & .362\\ 
\textbf{UGT (Ours)}                & \textbf{.850}      & \textbf{.620} & \textbf{.570} & \textbf{.501}  & \textbf{.875}& \textbf{.624}& \textbf{.565}  & \textbf{.511} & \textbf{.856}& .600& \textbf{.528}  & \textbf{.465}\\
\hline
& \multicolumn{4}{c}{\textbf{Texture}} & \multicolumn{4}{c}{\textbf{Pose}} & \multicolumn{4}{c}{\textbf{Shape}} \\
\hline
\textbf{CDAN~\cite{cdan18}}\tnote{**} & .820  & .532            & .420            & .364            & .844           & .620           & .521             & .450            & .773           & .561           & .491             & .441\\
\textbf{BSP~\cite{bsp19}}\tnote{**}   & .696  & .444            & .384            & .315            & .831           & .610           & .510             & .423            & .757           & .535           & .485             & .434\\
\textbf{MDD~\cite{mdd19}}\tnote{**}   & .895   & .518            & .427            & .400            & .870           & .611           & .534             & .469            & .836           & .541           & .459             & .386\\
\textbf{MCD~\cite{mmcd18}}\tnote{**}  & .896   & .522            & .432            & .392            & .865           & .623           & .532             & .471            & .834           & .538           & .456             & .397\\
\textbf{MCC~\cite{mcc19}}\tnote{**}   & .874   & .671            & .547            & .495            & .867           & .611           & .521             & .460            & .818           & .601           & .524             & .460\\ 
\textbf{FixBi~\cite{na2021fixbi}}\tnote{**}     & .854      &    .574       &  .445         & .369        &  .842   & .533 & .472 & .446   &   .801         & .500           & .435             & .373 \\
\textbf{MIC~\cite{hoyer2023mic}}\tnote{**}      &  .821     & .706          & .631          & .576        &  .799   & .613 & .509 & .455   &    .807        &    .608        &   .565           &.467  \\
\textbf{ToAlign~\cite{wei2021toalign}}\tnote{**}&  .594     & .413          & .312          & .273        &  .788   & .574 & .503 & .418   &   .719         & .548           & .460             & .391\\
\textbf{CST}~\cite{liu2021cyclecst}\tnote{**}   &   .858    &    .657       &   .538        &   .477      &  .887   & .617 & .525 &  .451  & .831           & .617           &   .495           & .441 \\
\textbf{DUA}~\cite{mirza2022normdua}\tnote{**}  & .918      &  .691         & .514          &  .468       &  .755   & .511 & .423 & .355   &  .695          &  .455          & .386             & .345\\
\textbf{DINE}~\cite{liang2022dine}\tnote{**}    &  .911     &  .572         &  .432         &  .401       & .885    & .618 & .543 & .448   &   .838         &  .520          &    .426          & .360\\
\hdashline
\textbf{RPL~\cite{rpl}}                      & .703               & .371          & .238          & .227          & .730         & .493         & .400             & .329            & .670           & .426           & .340             & .311\\       
\textbf{BNA~\cite{bna}}                      & .701               & .383          & .247          &.239           & .737          & .510          & .407         & .355           & .662            & .436           & .350           & .311         \\
\hdashline
\textbf{CompNet~\cite{compnet2020journal}}   & .747               & .539          & .462          & .426          & .768           & .581           & .538             & .458            & .698           & .466           & .451             & .400\\ 
\textbf{UGT (Ours)}                  & \textbf{.936}      & \textbf{.726} & \textbf{.665} & \textbf{.635}  & \textbf{.892}& \textbf{.632}& \textbf{.555}  & \textbf{.481} & \textbf{.852}& \textbf{.644}& \textbf{.601}  & \textbf{.567}\\
\hline
\end{tabular}
\begin{tablenotes}
\item[**] \footnotesize{\textbf{Pretrained Imagenet Backbone used (Resnet-50) / Pretrained UDA model used}.}
\end{tablenotes}
\end{threeparttable}
\end{table*}
\end{tiny}
}

\newcommand{\ablationtottexext}{
\begin{tiny}
\begin{table*}
\vspace{-2mm}
\begin{center}
\caption{Ablation analysis for (a) OOD-CV~\cite{zhao2021robin} Combined (b) OOD-CV Texture (c) Imagenet-C (Snow) Corruption}
\label{table:ablation-ext}
\begin{tabular}{lcccc|cccc|cccc}
\hline\noalign{\smallskip}
 Occlusion$\rightarrow$ &       L0      & L1 &  L2 &  L3 &   L0    & L1 &  L2 &  L3& L0  & L1 &  L2 &  L3\\
\noalign{\smallskip}
\hline
\noalign{\smallskip}
\textbf{Baseline(B)}       & .698           & .466           & .451             & .400 & .715 &.575 &.475& .409 & .529 &.348 &.258 &.210\\   
\textbf{\hspace{.25cm}$+\Lambda^{\mathcal{R}}+\mathcal{A^R}$}    &  .816 & .598  & .524 &  .498 &.785 &.660 &.559& .515&.781 &.671 &.582 &.480\\
\textbf{\hspace{.25cm}$+\Lambda^{\mathcal{R}}+\mathcal{A}^{\mathcal{R'}}$}    & .852 &  .644 & .601 & .567 & .843& .764& .656 & .623 &.885 &.742 &.634 &.523\\
\hline
\end{tabular}
\end{center}
\vspace*{-8mm}
\end{table*}
\end{tiny}}

\definecolor{cvprblue}{rgb}{0.21,0.49,0.74}
\usepackage[pagebackref,breaklinks,colorlinks,citecolor=cvprblue]{hyperref}
\usepackage{arydshln}
\usepackage{threeparttable}

\def\paperID{8546} 
\def\confName{CVPR}
\def\confYear{2024}

\title{A Bayesian Approach to OOD Robustness in Image Classification}

\author{Prakhar Kaushik\\
Johns Hopkins University\\
{\tt\small pkaushi1@jh.edu}
\and
Adam Kortylewski\\
University of Freiburg\\
{\tt\small akortyle@mpi-inf.mpg.de}
\and
Alan Yuille\\
Johns Hopkins University\\
{\tt\small ayuille1@jh.edu}
}

\begin{document}
\maketitle
\begin{abstract}
An important and unsolved problem in computer vision is to ensure that the algorithms are robust to changes in image domains. 
We address this problem in the scenario where we 
only have access to images from the target domains. Motivated by the challenges of the OOD-CV~\cite{zhao2021robin} benchmark where we encounter real world Out-of-Domain (OOD) nuisances and occlusion, we introduce a novel Bayesian approach to OOD robustness for object classification. Our work extends Compositional Neural Networks (CompNets), which have been shown to be robust to occlusion but degrade badly when tested on OOD data. We exploit the fact that CompNets contain a generative head defined over feature vectors represented by von Mises-Fisher (vMF) kernels, which correspond roughly to object parts, and can be learned without supervision. We obverse that some vMF kernels are similar between different domains, while others are not. This enables us to learn a transitional dictionary of vMF kernels that are intermediate between the source and target domains and train the generative model on this dictionary using the annotations on the source domain, followed by iterative refinement. 
 \alan{Our work extends Compositional neural networks (CompNets) which have been shown to be robust to occlusion but degrade badly when tested on OOD. We exploit the fact that CompNets contain a generative head defined over feature vectors represented by vMF kernels, which correspond roughly to object parts, and can be learned without supervision. We observe that some vMF kernels are similar between different domains, which others are not. This enables us to learn a transitional dictionary of vMF kernels that are intermediate between the source and target domains and train the generative model on this dictionary using the annotations on the source domain, followed by iterative refinement. We show that this
 UGT (SPELL IT OUT!!) which uses the CompNet architecture performs very well on ODD even when occlusion is added.}
This approach, termed Unsupervised Generative Transition (UGT), performs very well in OOD scenarios even when occlusion is present. UGT is evaluated on different OOD benchmarks including the OOD-CV dataset, several popular datasets (e.g., ImageNet-C~\cite{imagenetC}), artificial image corruptions (including adding occluders), and synthetic-to-real domain transfer, and does well in all scenarios. 
\end{abstract}
\newcommand\blfootnote[1]{%
  \begingroup
  \renewcommand\thefootnote{}\footnote{#1}%
  \addtocounter{footnote}{-1}%
  \endgroup
}
\section{Introduction}
\label{sec:intro}
 In\blfootnote{This work has been supported by Army Research Laboratory award W911NF2320008 and ONR with N00014-21-1-2812. A Kortylewski acknowledges support via his Emmy Noether Research Group funded by the German Science Foundation (DFG) under Grant No. 468670075.} recent years, machine learning algorithms have been extremely successful for tasks like object classification when evaluated on benchmarked datasets like ImageNet. But these successes require that the training and test data (or the source domain and the target domain data) be identically and independently distributed (IID) from some underlying source. However, in practice, it is important to ensure that the algorithms generalize to data that differ from the training data. 
 For example, in real-world applications, an algorithm for car detection may encounter cars with unusual shapes and textures (\cref{fig:occrobin}), which did not occur in the training set. 


\begin{figure*}[ht]
\centering
\includegraphics[width=0.9\linewidth]{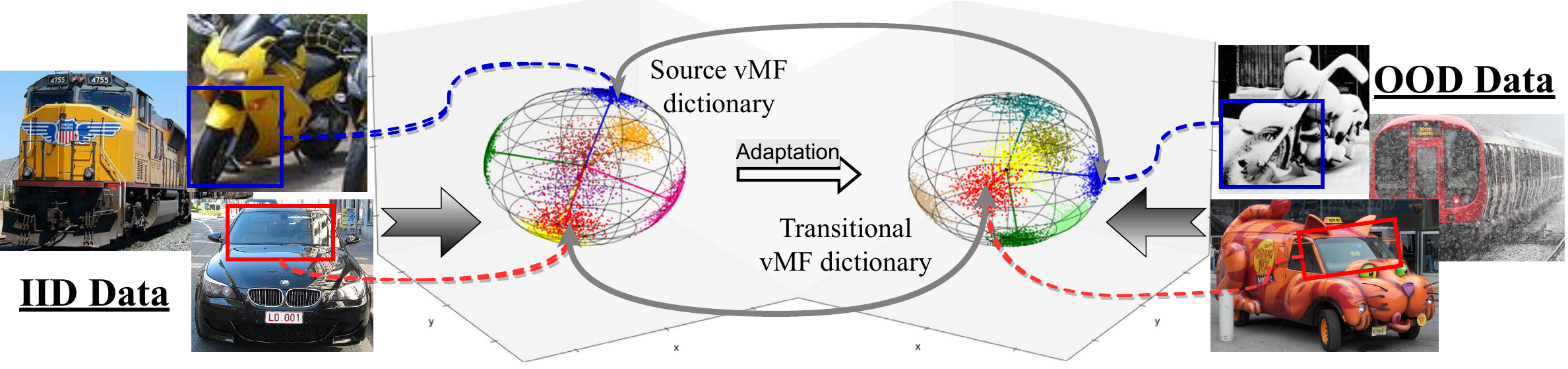}
\caption{Illustration of the key principle underlying our Bayesian approach. Related work has shown that clusters of feature vectors learned in an unsupervised manner resemble part-like patterns \cite{vc17, compnet2020journal}. We observe that some feature clusters (represented here on a vMF manifold) are very similar in both IID and OOD data (illustrated in blue and red boxes), whereas for other feature clusters there is no corresponding equivalent in the other domain. Our Bayesian approach exploits this property by first learning a generative model of feature clusters and their spatial combinations on the IID data and subsequently adapting the model to OOD data via an unsupervised adaptation of the vMF cluster dictionary, while retaining the spatial relations between clusters.}
\label{fig:transitional}
\end{figure*}


Existing OOD methods~\cite{imagenetA,imagenetC,imagenetR,pretranshend21,recht2019imagenet} have shown success in dealing with robustness issues when evaluated on early robustness datasets, such as Imagenet-C~\cite{imagenetC}, Imagenet-R~\cite{imagenetR}, and Imagenet-A~\cite{imagenetA}, where the domain differences are due to synthetic corruptions, adversarial images, rendered images, and similar factors~\cite{zhao2021robin}. But these algorithms performed less well on a newer benchmark, OOD-CV~\cite{zhao2021robin}, which focuses on systematic analysis of real-world nuisances, e.g. changes in texture, 3D pose, weather, shape, and context. From a related perspective, OOD-CV studies the causal factors that result in the domain gap~\cite{DBLP:journals/corr/abs-2112-00639}. In addition, previous works have rarely been evaluated for robustness to occlusion, an important OOD robustness metric. 

In this work, we address OOD robustness on OOD-CV, and related datasets, focusing on real-world domain differences and occlusion.  We build on a class of Bayesian neural models called Compositional Neural Networks (CompNets), as they have been shown to be robust to partial occlusion~\cite{Kortylewski2020CompositionalCN,compnet2020journal,sun2020weakly,Yuan_2021_CVPR}. This is achieved by replacing the discriminative head of a CNN with a generative model of the feature vectors based on the objects' spatial geometry. However, CompNets are fully supervised and are not robust to OOD nuisances. In this work, we develop an unsupervised approach, Unsupervised Generative Transition (\textbf{UGT}), which generalizes CompNets to OOD scenarios.

UGT relies on intuition that in OOD scenarios, the appearance of object parts is highly variable (due to changes like texture or weather), while the spatial geometry of objects is often fairly similar between domains. We analyze CompNets and modify them to take advantage of the intuition mentioned above. By introducing a {\it transitional dictionary} of von Mises-Fisher~\cite{kitagawa2022von} kernels (\cref{fig:transitional}), which shares the properties of both domains, we can intuitively learn the spatial geometry of the source and transfer it to the target domain. 
UGT leverages the property that the hierarchical structure of generative models like CompNets can be learned in a two-stage manner. 
\textbf{1)} An unsupervised learning stage of a dictionary of neural network features, called {\it vMF kernels}, using clustering in both source and target domains. 
The vMF kernels intuitively represent local object part structures. 
\textbf{2)} A supervised learning stage of the spatial relations of the vMF kernels on the source domain. 
\alan{NOTE -- This paragrph can be removed if necessary. More precisely, we exploit a hitherto unexploited property of CompNets which is that their vMF kernels can be learned without supervision although learning the generative head requires supervision. This enables us to learn a {\it transitional vMF kernel dictionary} which contains properties of both the source and target domains. We train the generative model for the CompNets on the annotated source data, using the transitional vMF kernels, and show that this model performs well on the target domain and improves with pseudo-labeling. A crude intuition for this approach is that the vMF kernels capture the visual appearance of parts which can differ between the source and target domain. We observe that some vMF kernels are very similar in both domains and, for example, vehicle windows largely remain unchanged even when the context, texture and shape of the vehicle changes (\cref{fig:three graphs}). This motivates us to learn a {\it transitional} dictionary which is learned on the target domain, but where the feature clusters are encouraged to be similar to those on the source domain. By contrast, the parameters of the generative model capture the spatial structure of the object (e.g., which parts are likely to appear in specific locations), hence it is fairly similar in both source and target domain, and can be learned using only supervision on the source domain.}\\
We primarily evaluate UGT on the OOD-CV benchmark~\cite{zhao2021robin}. In addition, to challenge UGT, we add occluders to OOD-CV and create a new dataset called \textit{Occluded-OOD-CV} (\cref{o-oodcv}). We also test UGT on Imagenet-C corruptions and Synthetic-to-Real domain robustness.\alan{Prakhar needs to edit this.} Our studies show that UGT performs well on all these tasks and significantly outperforms the SOTA baselines. 

We make several important contributions in this paper.
\begin{enumerate}[topsep=1pt,itemsep=1pt,partopsep=1pt, parsep=1pt]
    \item We model objects by a generative model on feature vectors. 
    Our method, UGT, extends CompNets~\cite{compnet2020journal} by decoupling the learning into unsupervised learning of vMF kernels and supervised learning of the spatial geometry enabling us to learn transitional dictionaries.
    \item UGT achieves state-of-the-art results on the real-world OOD robustness problem on the OOD-CV dataset~\cite{zhao2021robin} and demonstrates exceptional performance on generalizing under the synthetic corruptions of Imagenet-C.
    \item UGT also achieves strong results for the Synthetic-to-Real scenario (UDAParts~\cite{udapart22} to Pascal3d+) dataset. 
    \item We introduce the Occluded-OOD-CV dataset by adding occluders to OOD-CV and show that UGT is robust to this compounded problem of occlusion and nuisance.
\end{enumerate}
\begin{figure*}[th]
\centering
\includegraphics[width=0.85\linewidth]{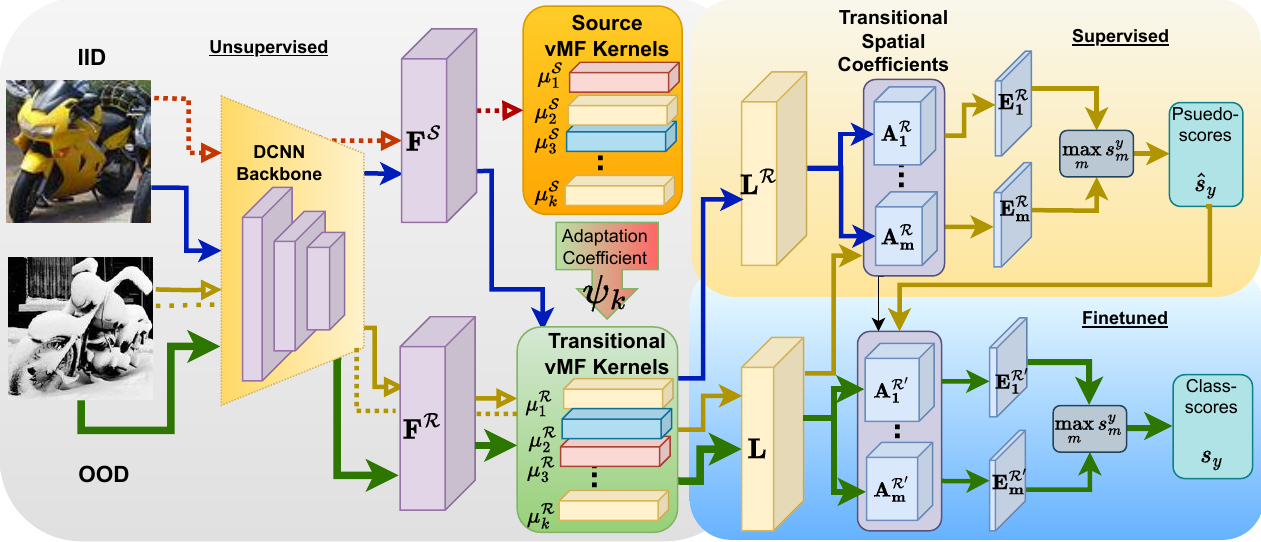}
\caption{Rough illustration of our Bayesian method. (\scalebox{1.5}{{\textcolor{myred}{$\dashrightarrow$}}},\scalebox{1.5}{\textcolor{myyellow}{$\dashrightarrow$}}) A DCNN backbone is used to extract the source (IID) $F^\mathcal{S}$ and target (OOD) features
$F^\mathcal{R}$. The source feature vectors $F^\mathcal{S}$ are then used to learn the source vMF kernels that are then adapted to the transitional vMF kernels using target domain features $F^\mathcal{R}$ and the adaptation coefficients $\psi_k$ in an unsupervised manner. (\scalebox{2}{\textcolor{myblue}{$\rightarrow$}}) Transitional Spatial coefficients ($A^{\mathcal{R}}$) are then learned using the transitional vMF likelihood $L^\mathcal{R}$ i.e. non-linear activation applied to a convolution of $F^\mathcal{S}$ and transitional kernels using source labels. (\scalebox{2}{\textcolor{myyellow}{$\rightarrow$}}) These spatial coefficients are then finetuned ($A^{\mathcal{R}'}$) using pseudo-scores \{$\hat{s}$\} generated using the transitional mixture likelihood $E^\mathcal{R}$ of target domain features $F^\mathcal{R}$. (\scalebox{2}{\textcolor{mygreen}{$\rightarrow$}}) shows the final feedforward pipeline during inference.\yaoyao{todo: We need to add the caption to describe the whole framework and make this figure self-contained.}}
\label{fig:algofig}
\end{figure*}
\section{Related Works}\label{sec:related} 
\yaoyao{This section needs to include two parts: 1) OOD robustness; 2) CompNets. In each part, we need to discuss: 1) recent articles on this topic; 2) how they relate to our work.}
OOD robustness can be considered a subset of the larger unsupervised domain adaptation problem and is closely related to domain generalization and transfer learning. 
Although related to both, our work focuses on OOD robustness. Our aim is to generalize well to an unlabelled target domain which is parameterized by real world nuisance factors like weather, shape, pose, texture changes and partial occlusion - which often leads to drastic changes to visual scenes and objects
not found in the source dataset.

In the past few years, there has been an increase in the number of works~\cite{imagenetA,imagenetC,imagenetR,pretranshend21,recht2019imagenet} that characterize model performance on OOD data and treat this as a measure of robustness. 
The common idea that underlies most works is to leverage a property of the unlabeled target domain to allow generalization of a model trained on the source domain. There have been successful efforts to use feature statistics to adapt to the new domain; e.g., Sun et al.~\cite{coral} try to minimize domain shift by aligning the second-order statistics of source and target distributions; Bug et al.~\cite{fbnorm17} employ feature aware normalization with gating elements from Long Short-Term Memory units for normalization among different spatial regions of interest. Some methods employ techniques based on adaptive batch normalization and weight normalisation~\cite{bna}. Other methods include self-learning using entropy minimization~\cite{wang2020tent}, adaptive pseudo-labeling techniques~\cite{song2020learning,NIPS2016_b59c67bf,Lee_pseudo-label:the,galstyan2007empirical} and robust lost functions~\cite{gce,ghosh2017robust}. 
\alan{This section should be simplified. Surely its only purpose it to explain why we use OOD-CV with occlusion instead of only using OOD-CV? This can be said more simply. We have already explained, in the last section, why we are using OOD-CV. So here we simply need to explain why are also testing on OOD-CV with occlusion. Most of the first paragraph is redundant and can be cut.}
\prakhar{removed section and 1st paragraph}
Although, current works have been successful at dealing with robustness problems when evaluated on earlier robustness datasets \cite{imagenetA,imagenetC,imagenetR} they have been shown to struggle with real world nuisances (OOD-CV \cite{zhao2021robin}) and occlusion \cite{compnet2020journal,kaushik2024sourcefree}. Few generative Bayesian methods such as CompNets \cite{compnet2020journal, sun2020weakly, vc17} have shown their relative robustness to occlusion, but still struggle with other OOD nuisances.
\section{Method}
\noindent
We address OOD robustness from a Bayesian perspective which, to the best of our knowledge, is novel. Our starting point is a class of generative models, described in \cref{sec:simple}, which have been shown to be robust to occlusion \cite{compnet2020journal} when not dealing with other OOD nuisances. 
We describe method motivation 
in \cref{ourmethod} and the technical details in \cref{sec_details}.
\subsection{Bayesian Neural Architecture}\label{sec:simple}
Our base architecture is similar to CompNets~\cite{compnet2020journal} and is explained in this section to help readers unfamiliar with them. Our method extends this class of neural models by non-trivially modifying the training methodology to enable OOD robustness along with occlusion robustness.

This class of models differs from conventional Deep Networks by replacing the discriminative head 
by a generative model of feature vectors. 
For each object $y$ we learn a generative model $P(F|y)$ for the feature vectors $F$. This model is formulated as a mixture model $P(F|y) = \sum _m P(F|y,m)$ where the mixture variable $m$ roughly corresponds to the viewpoint of the object. The conditional distributions $P(F|y,m)$ for the features are factorizable in terms of position so that $P(F|y,m) = \prod _{a \in \mathcal{D}} P(f_a|y,m)$, where $a \in \mathcal{D}$ specifies the position in the image. These distributions $P(f_a|y,m)$ are specified in terms of {\it von Mises-Fisher (vMF) dictionaries}, with parameters $\Lambda = \{\sigma _k, \mu _k\}$ and by {\it spatial coefficients} with parameters $\mathcal{A}= \{\alpha_{a,k}^{y,m}\}$. 
We use the following generative probability distribution for the neural features $F$ conditioned on an object $y$~\cite{Kortylewski2020CompositionalCN,compnet2020journal}:
\noindent
{\small
\begin{align} 
&P(F|y) \hspace{-0.04cm} = \hspace{-0.08cm} \medmath{\sum_m} P(F|y,m) \hspace{-0.04cm} = \hspace{-0.08cm} \medmath{\sum_m \hspace{-0.06cm} \prod_{a \in \mathcal{D}}} \hspace{-0.07cm} P_a(f_a|y,m) P(m)\label{eq:1},\\
&P_a(f_a|y,m)\hspace{-0.04cm}=\hspace{-0.04cm}P_a(\medmath{f_a|\mathcal{A},\Lambda}) \hspace{-0.04cm}=\hspace{-0.04cm}\sum_k \alpha_{a,k}^{y,m} P(f_a|\sigma_k, \mu _k)\label{eq:2},\\
&P(f|\sigma_k, \mu _k) = \frac{e^{\sigma_k \mu_k^T f}}{\medmath{Z(\sigma_k)}}, ||f|| = 1, ||\mu_k|| = 1\label{eq:3},
\vspace{-4mm}
\end{align}}%
\noindent 
We typically use 4 mixture components in our method and $P(m)$ is an uniform prior over the mixture components.
As shown in \cite{vc17, compnet2020journal} \alan{Is it? Or do we go back to the original visual concept papers?} each vMF kernel can be qualitatively interpreted as a subpart of the object (i.e., all image patches with feature responses close to $\mu _k$ look like visually similar object subparts). We use von Mises-Fisher distributions instead of Gaussian distributions because the feature vectors $f_a$ and the means $\mu_k$ must have a unit norm \cite{hasnat2017mises,pmlr-v32-gopal14}.
The {\it spatial coefficients} $\mathcal{A}= \{\alpha_{a,k}^{y,m}\}$ specify the probability that the vMF kernel $k$ occurs at the position $a$ conditioned on the object $y$ and its mixture component $m$.

\textbf{Inference.} After learning, inference on an image with feature vectors $F$ is performed by a forward pass which estimates which object is more likely to generate the features $F$ of the input image, ${\hat y} = argmax_y P(F|y)$~\cite{compnet2020journal,sun2020weakly}. 

\textbf{Occlusion modeling.}\label{occmodel} 
To make the model, described above, robust to occlusion (in non-OOD data), an outlier process is added to allow for some of the image features to be generated by the object and others by a separate outlier process \cite{Kortylewski2020CompositionalCN,sun2020weakly}. This is formalised by:
\vspace{-1mm}
{\small
\begin{equation}
\label{eq:occ}
P(\medmath{F|y})=\hspace{-.1cm} \medmath{\prod _{a \in \mathcal{D}} \sum_m} P_a(\medmath{f_a|y,m}) ^{z_a}Q(\medmath{f_a})^{1 - z_a}P(\medmath{m})P(\medmath{z_a})
\vspace{-2mm}
\end{equation}
}%
\noindent 
where $Q(f_a)$ is a vMF distribution for a feature generated by an occluder which can be estimated from non-annotated images~\cite{kortylewski2018empirically,Yuan_2021_CVPR,compnet2020journal}. The latent variable $z_a \in \{0,1\}$ indicates whether pixel $a$ is occluded or not occluded ($z_a=\{1, 0\}$ respectively) and the prior $P(z_a)$ indicates the prior probability of a pixel being occluded. Note that we could also, in theory, sum over $z$ (we currently take a max).

\textbf{Training} CompNets~\cite{compnet2020journal,Yuan_2021_CVPR,sun2020weakly} are trained end-to-end to optimize the model parameters $\Lambda, \mathcal{A}$ using the standard supervision for object classification (e.g., the mixture components and the vMF kernels are treated as latent variables). In an OOD scenario the image features no longer correspond well with the learned generative model and without labels, we cannot trivially finetune the model. UGT utilizes an insightful training strategy to solve this problem.
\begin{algorithm*}
\caption{Unsupervised Generative Transition}\label{alg:cfm}
\begin{algorithmic}[1]
\State \textbf{Input:} Set of source domain images $I^{\mathcal{S}} = \{I^{\mathcal{S}}_1,...,I^{\mathcal{S}}_n\}$, target domain images $I^T=\{I^T_1,..., I^T_N\}$, source domain labels $y=\{y^{\mathcal{S}}_1,...,y^{\mathcal{S}}_n\}$, deep network backbone $\Gamma(.,\zeta)$, background images $\mathcal{B}_{i=1}^r$
\State \textbf{Output:} Target domain model parameters $\mathcal{T}=(\mathcal{A},\Lambda)$, background model $\beta_r$
\Procedure{UGT}{$I^{\mathcal{S}}, I^T, y, \Gamma, \beta_r $}
\State $\{F^{\mathcal{S}}\}, \{F^\mathcal{R}\} \longleftarrow \Gamma((\{I^{\mathcal{S}}\}, \{I^T\}), \zeta)$ \Comment{\parbox[t]{.5\linewidth}{Extract source \& target featuremaps from DCNN backbone}}
\State $\Lambda^{\mathcal{S}}(\mu_k)\longleftarrow \text{cluster \& \small {MLE}(\{$F^{\mathcal{S}}$\})}$ \Comment{\parbox[t]{.5\linewidth}{Initialize source vMF kernels by kmeans \& learn using MLE}}
\State $\Lambda^{\mathcal{R}}_{initial}(\mu)\longleftarrow \Lambda^{\mathcal{S}}(\mu_k)$ \Comment{\parbox[t]{.5\linewidth}{Initialise transitional vMF kernels with source vMF kernels}}
\State $\Lambda^{\mathcal{R}}(\mu)\longleftarrow \text{MLE}(F^T, \Delta(\psi,\Lambda^{\mathcal{S}},\Lambda^{\mathcal{R}}))$ \Comment{\parbox[t]{.5\linewidth}{Learn transitional vMF features using regularized MLE with target domain data (\cref{transdict}, \cref{eq:5}-\cref{eq:9})}}
\State $\{L^{\mathcal{R}}\}\longleftarrow \sum _k \pi _k e^{\sigma _k \mu _k^T f^{\mathcal{S}}}/Z[\sigma _k](F*\Lambda^{\mathcal{R}}(\mu_k))$ \Comment{\parbox[t]{.5\linewidth}{Compute regularized transitional vMF likelihood with source featuremaps and transitional vMF kernels}}
\State $\mathcal{A^R}_{y_s,m} \longleftarrow$\text{ cluster\&\small {MLE}}$(\{L^{\mathcal{R}}\},y_S)$ \Comment{\parbox[t]{.5\linewidth}{Calculate spatial coefficients using transitional vMF likelihood and source feature vectors (\cref{transpat})}}
\State $y_{\hat{T}}\longleftarrow argmax_y P(F|\Lambda^{\mathcal{R}}, \mathcal{A^R})$\Comment{\parbox[t]{.5\linewidth}{Pseudo-label target domain data using transitional model }}
\State $\mathcal{A^{R'}}_{y_{\hat{T},m}}\longleftarrow\text{cluster}\&\text{\small {MLE}}(\{L^{\mathcal{R}}\},y_{\hat{T}})$ \Comment{\parbox[t]{.5\linewidth}{Finetune spatial coefficients using pseudolabelled data $y_{\hat{T}}$}}
\State $\mathcal{T}\longleftarrow optimize(\mathcal{L}_{\text{\tiny gce}} + \psi_v\mathcal{L}+ \psi_{\alpha}\mathcal{L})$ \Comment{\parbox[t]{.5\linewidth}{Optionally, finetune entire model using $y_{\hat{T}}$ (\cref{eq:12})}}
\EndProcedure
\end{algorithmic}
\end{algorithm*}
\subsection{Motivation on Generalizing to OOD Data}\label{ourmethod}
\noindent
UGT builds upon by the aforementioned Bayesian model 
because it gives a natural way to formulate an occluder process. These models, however, do not do well on OOD data (\cref{exps}). To solve this problem in an unsupervised manner requires reformulation of the training process. We motivate our solution for OOD, UGT, in following stages.

\textbf{Firstly}, the vMF kernel dictionaries (i.e., the subparts of the object) can be learnt without supervision and hence can be found on both the source (annotated) and the target (non-annotated) domains. \textbf{Secondly}, we observe that some of the vMF kernels are similar between different domains (intuitively some subparts are similar between both domains). \textbf{Thirdly}, we can build on this observation to learn a transitional dictionary, which encourages vMF kernels in both domains to be similar if possible\alan{come back to this later}, and which works well in both domains. \textbf{Fourthly}, we note that the spatial coefficients  capture the spatial activity pattern of the vMF kernels and these patterns depend on the spatial structure of the objects and so are mostly invariant to the domain, which suggests that we can learn the spatial coefficient on the source domain (where annotations are available), provided we use the transitional dictionary of vMF kernels, and that these spatial coefficients give a good initial estimate for the spatial coefficients on the target domain (which can be improved by simple pseudo-labeling). 

In the \textbf{first stage} we learn the vMF dictionaries $\Lambda$ without supervision by maximum likelihood estimation (MLE) assuming that the feature vectors $\{f_{a}\}$ of all the images (and at all positions) in each domain are generated by a mixture of von-Mises-Fisher distributions  $P(f|\Lambda) = \sum _k \pi _k e^{\sigma _k \mu _k^T f}/Z[\sigma _k]$. This is essentially clustering similar to that used in earlier studies \cite{vc17,compnet2020journal}. After the $\Lambda$ are learnt, if annotations are available (i.e., we know the object $y$) then we can learn the spatial coefficients $\mathcal{A}$  from the data $\{F_n\}$ in the annotated (source) domain by MLE from the distribution $\sum _m \prod _{a \in D} \sum _k \alpha ^{y,m}_{a,k} P(f_a|\sigma _k, \mu _k)$.

In the \textbf{second stage}, we compare the vMF dictionaries $(\Lambda^{\mathcal{S}})$ and $(\Lambda^T)$ on the source (S) and target (T) domain respectively. We observe that a subset of the dictionary vectors are similar, as measured by cosine similarity in the vMF feature space (\cref{fig:transitional}). \alan{Note: this is not illustrated conceptually in \cref{fig:three graphs} but maybe it should be?}We \textbf{conjecture} that this is because a subset of the vMF kernels, which correspond roughly to object subparts \cite{vc17}, is invariant to the nuisance variables which cause the differences between the domains. For example, for an object like a car or bus, some subparts like wheels and license plates may be very similar between the source and target domains but others may not (\cref{fig:transitional}). 

\noindent
These observations 
motivate us to learn a {\it transitional vMF dictionary} ($\Lambda^\mathcal{R}$). 
This dictionary is learnt by learning the dictionary on the target domain but adding a prior (or regularization constraint) that the dictionary elements in both domains are similar.  
Finally, we learn the spatial coefficients ${\bf \mathcal{A}}$ on the source domain, but using the transitional dictionary (\cref{transpat}). This allows us to utilize object geometry knowledge from the source domain in the target domain.\alan{Can we give an intuition why this works?} 
As we show in our experiments\alan{refer to the experimental section} and ablation (\cref{exps}, \cref{ablation}), this model already works well on the target domain and can be improved by pseudo-labelling techniques.
\subsection{Training UGT}\label{sec_details}
\noindent
Our Bayesian method, \textbf{UGT}, involves 3 steps - 
\textbf{1)} primarily, learning transitional dictionary $\Lambda^{\mathcal{R}}$, 
\textbf{2)} learning transitional spatial coefficients $\mathcal{A}^{\mathcal{R}}$ using $f^{\mathcal{S}}$ and $\Lambda^{\mathcal{R}}$, and lastly 
\textbf{3)} finetuning the transitional parameters $(\Lambda^{\mathcal{R}},\mathcal{A}^{\mathcal{R}})$ using simple pseudo-labelling. Refer to \cref{fig:algofig} for a simple illustration. \alan{Please refer to the figure}
\subsubsection{Learning Transitional Dictionary}\label{transdict} 
We initialize the \textit{transitional} von Mises-Fisher(vMF) dictionary vectors with the learnt source domain vMF dictionary vectors, i.e., $\Lambda^{\mathcal{R}}_{\text{\tiny initial}} = \Lambda^{S}$.  
The source domain vMF dictionaries i.e., $(\Lambda^{\mathcal{S}}(\mu,\sigma))$ are learnt from the features $f^{\mathcal{S}}$ in source domain by MLE as described in \cref{sec:simple} using the EM algorithm~\cite{vc17}.
We can learn the transitional vMF dictionary parameters $\Lambda^{\mathcal{R}}$ from the target domain feature vectors $f^{\mathcal{R}}$ through a few ways. We can maximize the regularized likelihood shown in \cref{eq:5} using the EM algorithm used to calculate the source domain parameters. \cref{eq:5} shows the Bayesian parameterization of our transitional model and can be seen as a penalized or regularized form of maximum likelihood estimation. We penalize the distance between the initialized transitional mean vectors (which are the source parameters) and the learnt ones. This regularization (like others) also helps in avoiding overfitting.
Since, we fix $\sigma_k$ as constant to reduce computation, the normalization term $Z(\sigma)$ reduces to a constant, and we can derive the penalized log-likelihood term as shown in \cref{eq:6}. $\psi$ is a adaptation parameter discussed later.
\alan{But we didn't do this in those papers -- we did it in the earlier VC papers.}
\prakhar{Fixed. removed $f^T$}
{\small
\begin{align}
& p(f^\mathcal{R}|\Lambda^{\mathcal{R}})=\prod_n \sum_k \alpha_{k} P(f_a|\sigma_k, \mu _k) \nonumber\label{eq:5}\\
& \qquad\qquad\qquad\qquad \exp{(-\psi_k\sum_k(||\mu_k-\mu_k^{\mathcal{S}}||))}\\
& l(\Lambda^{\mathcal{R}}) = \sum^{n} \log(\sum^k \pi_k \frac{e^{\sigma_k \mu_k^T f_i}}{Z(\sigma_k)}) - \psi_k\sum^{n}\sum^k(||\mu_k-\mu_k^{\mathcal{S}}||)\label{eq:6}\\
& \medmath{||f|| = 1, ||\mu_k|| = 1}, \sigma=1 \implies Z(\sigma)=const. \nonumber
\end{align}}%
The Expectation step for learning the transitional parameters is similar the source version. In the first step, we calculate the summary statistics for the transitional parameters using the new data. For posterior probability defined as 
{\small
\begin{equation}
P(k|f_i,\Lambda)= \frac{\pi_k p(f_i|\mu_k,\sigma_k)}{\sum^K \pi_k p(f_i|\mu_k,\sigma_k)}
\end{equation}}%
\noindent
for the $k^{th}$ mixture and where $p(f|\mu_k,\sigma_k)$ is defined in \cref{eq:3}, we update the mixture parameters in the maximization step in a regularized manner as follows,
{\small
\begin{align}
&  \hat{\pi}_k = \nu[\psi_k^{\pi} \frac{1}{n}\sum_{i=1}^n P(k|f_i,\Lambda) + (1-\psi^{\pi}_k)\pi_k^S]\label{eq:8}\\
&  \hat{\mu}_k = \psi_k^{\mu} \mathcal{E}_k + (1-\psi^{\mu}_k)\mu_k^S\label{eq:9}
\end{align}}%
\noindent
where, $\mathcal{E}_k$ is the first moment or mean of the $k^{th}$ mixture calculated on the new data, $\nu$ is a scaling parameter to ensure that $\sum_k\pi_k=1$ and $\psi_k$ is an adaptation coefficient which is defined for each parameter and mixture. It can be defined in a data dependent manner~\cite{REYNOLDS200019}, i.e., $\psi_k^{\mu,\pi} = (\frac{\omega_k}{P(k|f_i,\Lambda)}+1)^{-1}$ where 
$w_k$ is an empirically set hyperparameter which controls the adaptation emphasis between source and transitional parameters. Empirically, we observed that the adaptation coefficient is not very sensitive to changes to its value and therefore, we increase it monotonically during the EM iterations. A $\psi_{k}$ for a specific vMF kernel $\mu_k$ at time-step $t$ in $\Lambda^{\mathcal{R}}$ stabilizes if the change in its likelihood component is below a threshold value over the previous EM iteration step t-1 and then $\psi_{k}$ value. We find that only using the parameter update works well. For simpler datasets, even directly learning the transitional dictionary would suffice.
\subsubsection{Learning Transitional Spatial Coefficients}\label{transpat} After learning $\Lambda^{\mathcal{R}}$, we use it to estimate the transitional spatial coefficients $(\mathcal{A^R}(\alpha))$ using the labeled source domain features $f^{\mathcal{S}}$ (using MLE).
The spatial coefficients represent the expected activation of a calculated vMF kernel $\mu_k$ at a position $a$ in the feature map for a specific class $y$. 
\begin{equation} 
P_a(\medmath{f_a|y_s,m;\mathcal{A^R},\Lambda^R}) \hspace{-0.04cm}=\hspace{-0.04cm}\sum_k \alpha_{a,k}^{y_s,m} P(f_a|\Lambda^{\mathcal{R}}(\sigma_k, \mu _k))\label{eq:11}
\end{equation}
\noindent
We can leverage the learnt transitional vMF kernel dictionary $\Lambda^{\mathcal{R}}$ to learn spatial coefficients $\mathcal{A^R}(\alpha)$ which represent the spatial relationships of the vMF dictionary vectors over the source domain data $D_S$. As these spatial coefficients $\mathcal{A^R}$ are conditioned on $\Lambda^{\mathcal{R}}$, they also correspond to parts of target domain features even when they are learned using $f^{\mathcal{S}}$, thus creating a transitional model with parameters ($\Lambda^{\mathcal{R}}, \mathcal{A^R}$) that we can use to classify target domain data.

This combination of conditioned transitional vMF dictionary $(\Lambda^{\mathcal{R}})$ and spatial coefficients $(\mathcal{A^R})$ can be leveraged to label a subset of target domain features, especially since we can focus on the subset of transitional vMF kernels $(\Lambda^{\mathcal{R}})$ which are similar to their source counterparts. We can use these pseudo labeled feature vectors $(y_{\hat{T}})$, along with $\Lambda^{\mathcal{R}}$ to finetune the current spatial coefficients $\mathcal{A^R}$ 
which leads to improved spatial coefficients $\mathcal{A^R}'$. 
\paragraph{Finetuning spatial coefficients.}\label{pseudo} 
 Transitional spatial coefficients ($\mathcal{A^R}$) are initialized with the values describing the expected activation of transitional vMF dictionary vectors $\Lambda^\mathcal{R}(\mu_k)$ for the source data features $f^{\mathcal{S}}$ at a position $a$ on a feature map $f_a$. Subsequently, we finetune these spatial coefficients $\mathcal{A^R}$ using a subset of target domain images that present high activations for the robust set of transitional vMF dictionary vectors $\Lambda^\mathcal{R}$. Optionally, we can also finetune $\Lambda^\mathcal{R}$ by relearning them without any initialization and regularization constraints. Although our model is trained by partitioning into two parts, it is still fully differentiable and trainable from end to end~\cite{Kortylewski2020CompositionalCN,Yuan_2021_CVPR,sun2020weakly}. We use this model property to finetune the entire model. The loss function (\cref{eq:12}) consists of a generalized cross entropy~\cite{gce} term calculated using the model predictions and two regularization parameters for the vMF dictionary and the spatial coefficient parameters. This is to encourage the vMF clusters to be similar to the feature vectors $f_a$. In \cref{eq:12}, $\zeta_{\{v,\alpha\}}$ represent the trade-off hyperparameters of the regularizing loss terms,
 \vspace{-1mm}
 \begin{equation}
     \mathcal{L} = \mathcal{L}_{\text{\tiny gce}}(y_{pred}, y_{\hat{T}}) + \zeta_v\mathcal{L}(F, \Lambda) + \zeta_{\alpha}\mathcal{L}(F, \mathcal{A}) \label{eq:12},
      \vspace{-1mm}
 \end{equation}
 For a constant vMF variance $\sigma_k$ (which also reduces the normalisation term to a constant) and assuming hard assignment of features $f_a$ to vMF dictionary clusters\cite{compnet2020journal},
 \begin{align}
     &\medmath{\mathcal{L}(F, \Lambda^\mathcal{R})=-\sum_a \max_k \log p(f_a|\Lambda^\mathcal{R}(\mu_k))}\label{eq:13}\\
     &\medmath{\mathcal{L}(F, \mathcal{A^R}')=-\sum_a (1-z_a)\log [\sum_k \alpha_{a, k}^{y_{\hat{T}},m} p(f_a|\Lambda^\mathcal{R}(\mu_k))]}\label{eq:14}
     \vspace{-2mm}
 \end{align}
Latent variable $z_a \in \{0,1\}$ is explained in \cref{occmodel}.

\section{Experiments}\label{exps}
\totalandudatable
Our experiments evaluate \textit{robustness} of vision classification models in an extended out-of-domain setup i.e., generalizing to target domains with individual nuisance factors and partial occlusion.
This allows us to thoroughly evaluate the efficacy of current methods which have been shown to perform well on other OOD robustness datasets on OOD-CV\cite{zhao2021robin} (which enables a systematic analysis of nuisances on real-world data), Occluded-OOD-CV (which allows us to evaluate models on a combination of partial occlusion with individual nuisances) and Imagenet-C corruptions (for analysis of synthetic corruptions). 
Lastly, we also show some initial results on Synthetic to Real OOD robustness using the UDAParts~\cite{udapart22} dataset.

\subsection{Setup and Data}
\paragraph{Datasets.} For primary evaluation, we use the OOD-CV~\cite{zhao2021robin} dataset. OOD-CV dataset consists of test subcategories which vary from the training data in terms of a main nuisance factor, namely, context, weather, texture, pose and shape. 
We use $L0$ for the $(0\%)$ occlusion level to represent this data setup in \cref{table:robintotaluda} and Supplementary Sec. B.

\begin{figure}[h]
    \centering
    \includegraphics[scale=0.38]{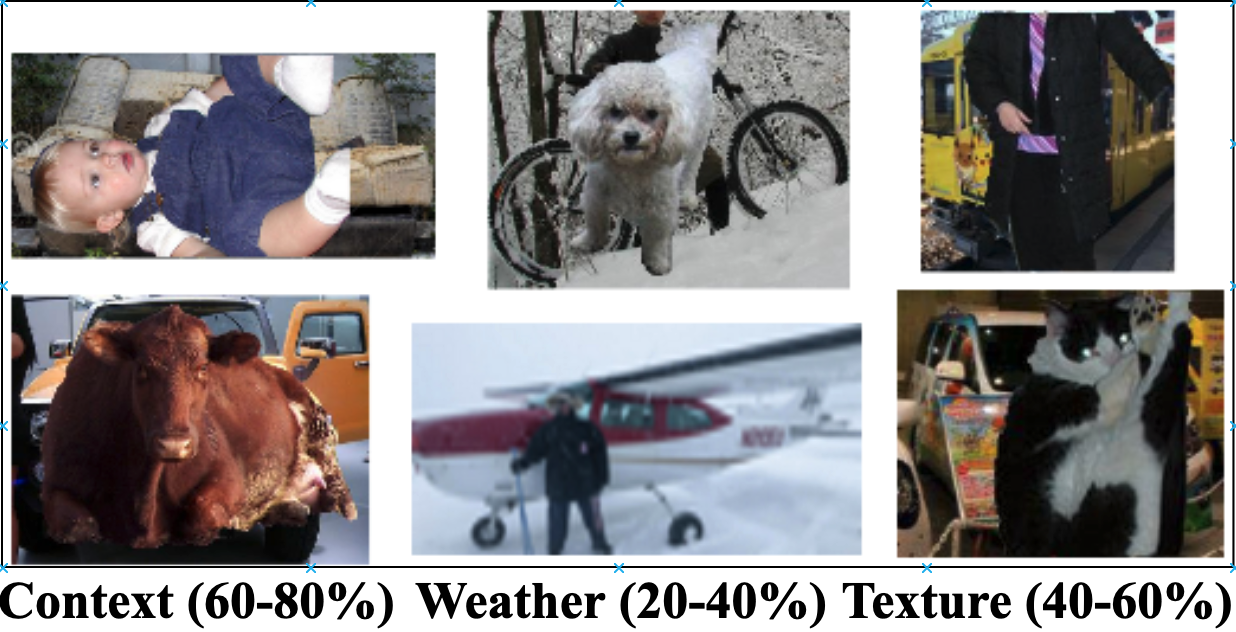}
    \caption{Occluded-OOD-CV dataset examples. Each object category is identified by its nuisance factor and occlusion percentage}
    \label{fig:occrobin}
\end{figure}

\textbf{Occluded-OOD-CV.}\label{o-oodcv} In addition to OOD-CV, we experiment with a more complex robustness analysis setup involving partial occlusion. In this setup, models that have been adapted in an unsupervised manner to target domains with nuisance factors are then evaluated on data with partial occlusion in addition to the real-world nuisances. For this purpose, we create a dataset named \textit{Occluded-OOD-CV} where we superimpose occluders on the OOD-CV test images objects in order to approximate real-life occlusion. These occluders have been cropped from the MS-COCO dataset, similar to \cite{Kortylewski2020CompositionalCN} and are superimposed on objects in the OOD-CV test set. There are three levels of partial occlusions - {\small $L1(20-40\%), L2(40-60\%) \text{ and } L3(60-80\%)$} which allows us to diversely analyze the occlusion robustness of the model (in addition to individual nuisance factors). \cref{fig:occrobin} shows some example images from our dataset. Previous works~\cite{pare21, compnet2020journal} have shown that using cropped occluders, as done in Occluded-OOD-CV, is akin to the use of real occluders for classification evaluation.
We also use Imagenet-C\cite{imagenetC} corruptions in the Pascal3D+ dataset for robustness evaluation with conventionally used synthetic corruptions. 
We also evaluate models in a synthetic (UDAParts~\cite{udapart22}) to real data (Pascal3D+~\cite{pascal3d14}) setup. 

In summary, we have $5$ different real world nuisance data subcategories (context, weather, texture, pose, shape), at least seven synthetic corruption categories (fog, pixelate, motion blur, etc.), one synthetic source dataset and $4$ partial occlusion levels (including no occlusion) for each experiment. We also run experiments on all the combined nuisance subcategories (\cref{table:robintotaluda}). So, in total we have $24$ sets of data and experiments for our (extended) OOD robustness setup on the OOD-CV dataset alone. 


\elasticgauss
\ablationtottexext
\textbf{Models.} We compare our work with our baseline method CompNets~\cite{compnet2020journal}, other well known recent works~\cite{rpl,bna} which have been shown to be SOTA on various \textit{robustness} datasets~\cite{imagenetC,imagenetA,imagenetR} as well as many well-known UDA methods~\cite{bnm20, mcc19, mdd19,cdan18,na2021fixbi,hoyer2023mic,wei2021toalign,liu2021cyclecst,mirza2022normdua,liang2022dine}. 
We focus on VGG16 and Resnet-50 backbones 
as they have been commonly used in most current methods\cite{Kortylewski2020CompositionalCN,bna,gce,rpl}.

\textbf{Training Setup.} All models are trained on the source data with corresponding labels. Models can access some unlabeled nuisance (target) data, which could be a single nuisance (OOD-CV, Imagenet-C), combined nuisances (\cref{table:robintotaluda}) or real data (when source data are synthetic). Models do not have access to images with partial occlusion at any time, and partially occluded images are only used for inference.
We also \textit{avoid} using different types of data augmentations and additional data training to have a fairer comparison amongst all the works. 
Although, our Bayesian model does not use \textbf{pretrained Imagenet backbones} for feature extraction for fairness, a number of our comparative methods~\cite{mcc19,mdd19,cdan18,bsp19,mirza2022normdua} 
perform poorly without one, so we relax this constraint for them. Our method is still capable of surpassing them in terms of classification accuracy. 

\noindent
Further details are provided in Supplementary Section C.
\subsection{Results}
\paragraph{OOD robustness to individual nuisances.} \cref{table:robintotaluda} (L0 columns) shows classification results on entire OOD-CV test data (combined nuisances) as well as five individual nuisances. We see that our model achieves state-of-the-art results in all experiments. In \cref{table:elasticgauss}, we observe that our model also performs exceedingly well when dealing with synthetic Imagenet-C corruptions. Refer to Supplementary Sec.C2 and Tables 5-11 for additional Imagenet-C results.
\paragraph{Synthetic to Real.} \cref{table:udaparts} 
shows our results on both normal and extended OOD robustness scenario in a synthetic to real setup, showing that our unsupervised method can robustly close the gap between its supervised counterpart while outperforming other methods by large margins.
\udaparttable
\paragraph{Extended OOD robustness under partial Occlusion.}
In \cref{table:robintotaluda}, \cref{table:elasticgauss} and Supplementary Tables 1-3, 5-11, our model outperforms other methods by significant margins in the \textit{extended OOD} scenarios of nuisance parameters with partial occlusion. We observe that the performance of other models which have been adapted to the target domain data drops drastically when encountering partial occlusion along with nuisance factors. This underlines the increased complexity of the extended OOD robustness scenario relative to the vanilla OOD robustness setup and how our Bayesian model is able to perform exceedingly well compared to conventional methods. 
\subsection{Ablation Analysis}\label{ablation}
\noindent
\cref{table:ablation-ext} and Supplementary Sec. D \& Tables 12-17 show the extensive results of the ablation study for UGT, underlying how each component contributes to the overall compositional model. 
We can see that just calculating the transitional vMF kernel dictionary ($\Lambda^{\mathcal{R}}$) and the transitional spatial coefficients $\mathcal{A}^{\mathcal{R}}$ 
improves the results significantly over the baseline method\cite{compnet2020journal}. Further finetuning the spatial coefficients
 ($\mathcal{A^R}'$) using pseudo-labelled target domain features 
 boosts the performance. We ablate our hypothesis regarding similar vMF kernels in source and target domains by visualizing image patches that are activated by similar cross-domain kernels (Supplementary Figures 9-11). We also ablate our hypothesis regarding robust spatial geometry by visualizing images activated by the same spatial coefficient in both source and target domains (using source and transitional vMF dictionaries) in Supp. Fig 4 and 7. Analysis of adaptation coefficient is discussed in Supp. Sec. E.
\section{Conclusion and Future Work}
In this work, we addressed the problem of developing object classification algorithms that are robust to OOD factors such as weather, context and occlusion. We generalize CompNets\cite{compnet2020journal} for OOD robustness by observing that they could be learned in two uncoupled steps: (i) unsupervised learning of a dictionary of vMF kernels (roughly corresponding to the subparts of the object) and (ii) supervised learning of the spatial structure of the objects (intuitively where the subparts occur). This enabled us to: (a) learn a transitional dictionary which captured the feature properties of both domains, and (b) learn the distribution of spatial structure on the source domain and transfer it to the target. This model is very successful and could be improved by simple pseudo-labeling techniques. Our empirical results on the OOD-CV\cite{zhao2021robin}, synthetic Imagenet-C corruptions, and the synthetic UDA-Parts dataset display the strong and versatile SOTA performance of our method. In addition, we developed a more challenging dataset Occluded-OOD-CV by introducing occlusion into OOD-CV and show that our Bayesian method, UGT, performed well in this difficult challenge. 
Our Bayesian approach could be extended to other tasks 
such as semantic segmentation, exploiting properties of CompNets\cite{occludedpascal, sun2020weakly} \alan{Refer to Angtian's CVPR paper and Yihong's AAAI paper}. We give a qualitative proof of concept in the Supplementary. 
{
    \small
    \bibliographystyle{ieeenat_fullname}
    \bibliography{egbib}
}
\end{document}


\title{A Bayesian Approach to OOD Robustness in Image Classification\\ Supplementary}
\maketitle
\tableofcontents
\begin{appendix}
\listoftables
\listoffigures
\end{appendix}


\thispagestyle{empty}
\begin{figure*}[!h]
    \centering
    \includegraphics[scale=.3]{figures/OOD sketch (2).png}
    \caption{Out of Domain Robustness for Object Classification using Unsupervised Generative Transition (UGT)}
    \label{fig:robinexample}
\end{figure*}
\begin{figure*}[!h]
    \centering
    \includegraphics[height=7cm]{figures/vmf new.png}
    \caption{Source and Transitional vMF dictionary Similarity}
    \subcaption{The cosine similarity between source $\Lambda^{\mathcal{S}}$ and transitional vMF dictionary $\Lambda^{\mathcal{R}}$ vectors (which are represented as circles and triangles respectively) in this conceptual vMF dictionary feature space is represented by the line connecting the circles and pentagons. The image patches from the source and target domains roughly corresponding to these vMF dictionary vectors are shown, which confirm that some similar image parts are represented by similar vMF dictionary vectors in both domains irrespective of the nuisance factor in the target domain. E.g. (lower right) image patches show windows from different vehicles - parts of objects which do not undergo much change when encountering nuisance factors like change in texture, shape and context of the vehicles.}
    \label{fig:three graphs}
\end{figure*}
\section*{List of Abbreviations}
\begin{abbrv}
\item[Avg.]                  Average
\item[plane]                 Aeroplane
\item[Occ.-OOD-CV]            Occluded-OOD-CV
\item[OOD-CV(L1)]                Occluded-OOD-CV (20-40\% occlusion)
\item[OOD-CV(L2)]                Occluded-OOD-CV (40-60\% occlusion)
\item[OOD-CV(L3)]                Occluded-OOD-CV (60-80\% occlusion)
\item[$\mathcal{A}$]         Spatial Coefficient
\item[$\Lambda$]             von Mises-Fisher (feature distribution) dictionary
\item[$\mathcal{S}$]         Source
\item[$\mathcal{R}$]         Transitional
\item[$\mathcal{T}$]         Target
\item[\textit{UGT}]          Unsupervised Generative Transition
\item[lr]                    Learning rate
\item[w.d]                   Weight decay
\item[$\kappa$]               vMF concentration parameter   
\end{abbrv}
\clearpage
\section{FAQs}
\begin{QandA}
   \item What are the differences/commonalities with Domain Adaptation and Domain Generalization w.r.t our domain robustness setup?
         \begin{answered}
         The problem of OOD robustness can be considered a niche subset of the larger unsupervised domain adaptation problem, and is closely related to domain generalisation and transfer learning as well. In unsupervised domain adaptation\cite{dasurvey21}, we often try to mitigate dataset biases between labelled source and unlabelled target data - there are often no constraints on type of domains and biases. The goal of domain generalisation is to learn a model, often from multiple related source domains, such that the resulting model can generalize well to any unseen target domain~\cite{Zhou_2022}.  Although related to both, our work focuses on an (extended) OOD robustness setup, where our aim is to generalize well to an unlabelled target domain which is parameterized by real world nuisance factors like weather, shape, pose, texture changes and partial occlusion - which often leads to drastic changes to visual scenes and objects in question not found in the source dataset.
         \end{answered}

   \item Difference in results from the original OOD-CV\cite{zhao2021robin} paper.
         \begin{answered}
            Differences, if any, in our experimental results may be attributed to different versions of same dataset used. In case of OOD-CV dataset\cite{zhao2021robin}, we contacted the authors and used their recommended version of the dataset. Additionally, we try not to use data augmentations
            and additional data to train our models which may also contribute to the difference. This is done to provide a fairer comparison between different methodologies. 
         \end{answered}
         
    \item What motivates our choice of datasets and models?
        \begin{answered}
        Our choice is motivated by the presence of real world nuisances and partial occlusion in the dataset. We chose datasets where either we could create partial occlusion due to presence of object masks or there was already a subset of the data which had partial occlusion. Our choice of datasets is also constrained by the amount of computational resources we use. As can be seen, we conduct a large number of experiments per model/architecture/dataset. For comparative models, we focus on works which have shown to work on previous robustness\cite{rpl, bna} benchmarks and some well-known UDA works\cite{cdan18, mdd19, mcc19, mmcd18, bsp19}. Secondary to out choice of real world nuisance factors and partial occlusion, we also test our method on Synthetic to Real transfer and synthetic corruptions in order to check its efficacy in these weel known setups.
        \end{answered}
         
    \item Applied partial occlusion is artificially applied on top of objects in images.
         \begin{answered}
         Although the occluded data that we use for evaluation for all our setups does have some naturally occluded images, we use real cropped occluders to ape partial occlusion in the real world. This might not be the best solution however it is the best available since there is a dearth of occlusion data for current vision models. Additionally, earlier works~\cite{occludedpascal, pare21} have shown that partial occlusion analysis for many computer vision tasks with such artificially placed occluders is similar to images with real partial occlusion.
         \end{answered}
         
    \item Can we sample from our generative model?
         \begin{answered}
         Our generative model defines a generative model on the level of intermediate neural features and not at the image level. Therefore, it is quite hard to visualize the sampled features.
         \end{answered}
    \item Is dictionary space misalignment a problem?
         \begin{answered}
         We empirically show via our SOTA results that dictionary space misalignment is not a big issue for our method. This is likely because the appearance of object parts may be highly variable, while the spatial geometry of the objects remains rather similar. 
         \end{answered}

    \item Is the target domain data accessible to models while training?
         \begin{answered}
         We follow the setup of OOD-CV\cite{zhao2021robin} paper. For Imagenet-C\cite{imagenetC} and UDAParts\cite{udapart22}, the target domain data available for finetuning is separate from the evaluation data.
         \end{answered}
    \item Rough intuition of our work.
        \begin{answered}
        in this work, we exploit a hitherto unexploited property of CompNets which is that their vMF kernels can be learnt without supervision although learning the generative head requires supervision. This enables us to learn a {\it transitional vMF kernel dictionary} which contains properties of both the source and target domains. We train the generative model for the CompNets on the annotated source data, using the transitional vMF kernels, and show that this model performs well on the target domain and improves with pseudo-labeling. A crude intuition for this approach is that the vMF kernels capture the visual appearance of parts which can differ between the source and target domain. We observe that some vMF kernels are very similar in both domains and, for example, vehicle windows largely remain unchanged even when the context, texture and shape of the vehicle changes (Figure~\ref{fig:three graphs}). This motivates us to learn a {\it transitional} dictionary which is learnt on the target domain but where the feature clusters are encouraged to be similar to those on the source domain. By contrast the parameters of the generative model capture the spatial structure of the object (e.g., which parts are likely to appear in specific locations), hence is fairly similar in both source and target domain, and be learnt using only supervision on the source domain.
        \end{answered}
\item Explanations regarding Equation 1-10.
         \begin{answered}
         Equation 1 is borrowed directly from earlier works\cite{compnet2020journal} where $P(m)$ is a uniform prior over the spatial mixture components and is written for completeness of the Bayesian parameterization. 
         
         In Equation 2 More mixtures may lead to better results as shown in previous works - we empirically find $4$ spatial mixtures to be sufficient for image classification task. 
         
         In Equation $4$, previous works recommend fixing the occlusion prior apriori $(0.5 - 0.6)$ given this probability can’t often be calculated. This prior is therefore largely uninformative, and we think calculating instance dependent occlusion prior is an interesting future work for this class of models. 
         
         In Equation 7, the prior probability is initialised using a vMF distribution with final source parameters and transitional features. The parameter update rules in Equation $9$ and Equation $10$ can be derived from the general MAP estimation equations for a vMF distribution as described in \cite{gauvin} (given for Gaussian Mixture Models, but applicable for vMF mixtures). Note that Equation 9 is an approximation of the derivation which empirically showed better results. The adaptation coefficients $\psi$ simply allow better tuning of adaptation rates between source and transitional data. Given that we use EM algorithm for MAP, we are guaranteed to converge to local minima under minor assumptions.
         \end{answered}
\end{QandA}

\section{Occluded-OOD-CV}

\begin{figure*}[!h]
    \centering
    \includegraphics[height=7.5cm]{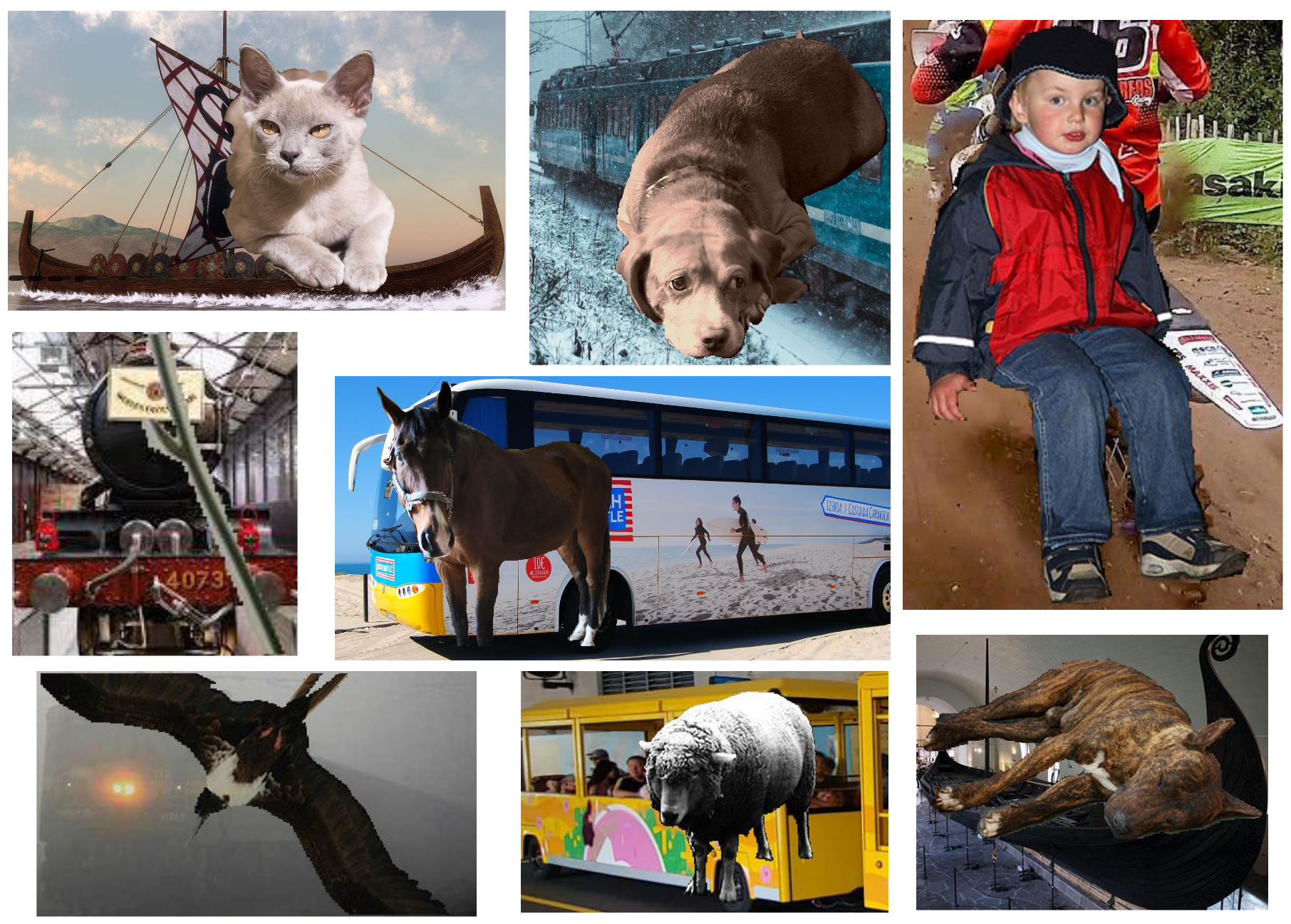}
    \caption{Examples from the Occluded-OOD-CV dataset}
    \label{fig:or-sup}
\end{figure*}

Occluded-OOD-CV is a dataset which is designed to evaluate the robustness of a model to partial occlusion in combination with a nuisance factor, namely, weather, texture, shape, 3D pose and context. In the Occluded-OOD-CV dataset, we simulate the partial occlusion by imposing samples of occluding objects on the primary image objects in OOD-CV. The superimposing occluders are either naturally present in the original OOD-CV test set or are cropped from the MS-COCO dataset\cite{mscoco}. We have $3$ levels of foreground occlusions in the Occluded-OOD-CV dataset - L1($20-40\%$), L2($40-60\%$) and L3($60-80\%$). The total number of (test) images for each level is 3820(L1), 3728(L2) and 3606(L3). A model which has been adapted to the OOD-CV nuisance data (in an unsupervised manner in our case) is then tested on the Occluded-OOD-CV data. The model is not shown the Occluded-OOD-CV data at any training or adapting stage. L0$(0\%)$ can be used to refer to the normal OOD-CV dataset. Figure~\ref{fig:or-sup} shows a few examples from the Occluded-OOD-CV dataset.

Given the dearth of real world occlusion datasets, we place the occluders \textit{artificially} on the image objects to ape real world occlusion. We believe that this a robust analysis of occlusion robustness as earlier works~\cite{occludedpascal, pare21} have shown that partial occlusion analysis for many computer vision tasks with such artificially placed occluders is similar to images with real partial occlusion.

\allbig
\contextbig
\weatherbig
\section{Experiments}
Our experiments are divided into three parts -
\begin{enumerate}
    \item \textit{Real World Nuisance Factors} - OOD-CV\cite{zhao2021robin} : Evaluation on this dataset is the focus of this work since it allows for real world (individual) nuisance robustness analysis. We can, therefore, test our models' capability to robustly adapt to nuisance-ridden real world data in an unsupervised manner.
    \item \textit{Synthetic Corruptions} - Imagenet-C\cite{imagenetC} : We also evaluate model's capability in adapting to synthetic corruptions, like, snow, gaussian blur, pixelate, elastic transform, etc, taken from Imagenet-C\cite{imagenetC}.
    \item \textit{Synthetic to Real} - UDAParts\cite{udapart22} (to Pascal3D+\cite{pascal3d14}) : We do an initial analysis of synthetic to real transfer using images created from UDAParts dataset. 
\end{enumerate}
Our choice of datasets is motivated by the fact that we want to evaluate a model's robustness to individual corruptions (real or synthetic) and partial occlusion. Since, our focus is primarily on evaluating model's robustness to unlabelled (real world) nuisance ridden data and occlusion, we choose OOD-CV\cite{zhao2021robin} as our primary evaluation data. We also apply Imagenet-C corruptions on Pascal3D+ objects\cite{pascal3d14} which allows us to evaluate our models on the compounded problem of synthetic corruptions and partial occlusion given that the dataset has a occluded version (Occluded Pascal)~\cite{occludedpascal}. 

In order to further evaluate the capabilities of our model, we also include a synthetic to real robustness analysis. For this, we use images rendered from the object models of the UDAParts~\cite{udapart22} dataset and then test our model on the same object set of Pascal3D+ dataset. Again, in inclusion to the synthetic data OOD robustness problem, we have partial occlusion added to the evaluation data~\cite{occludedpascal}.
\subsection{OOD-CV}
We evaluate all models on OOD-CV nuisance ridden test data (weather, context, texture, pose, shape) as well as our extended OOD setup which included Occluded OOD-CV with 3 levels of partial occlusions $(L1(0-20\%), L2(20-40\%), L3(40-60\%))$. The unoccluded data can also be referred to as L0.

\begin{figure*}[!htb]
    \centering
    \includegraphics[height=10cm]{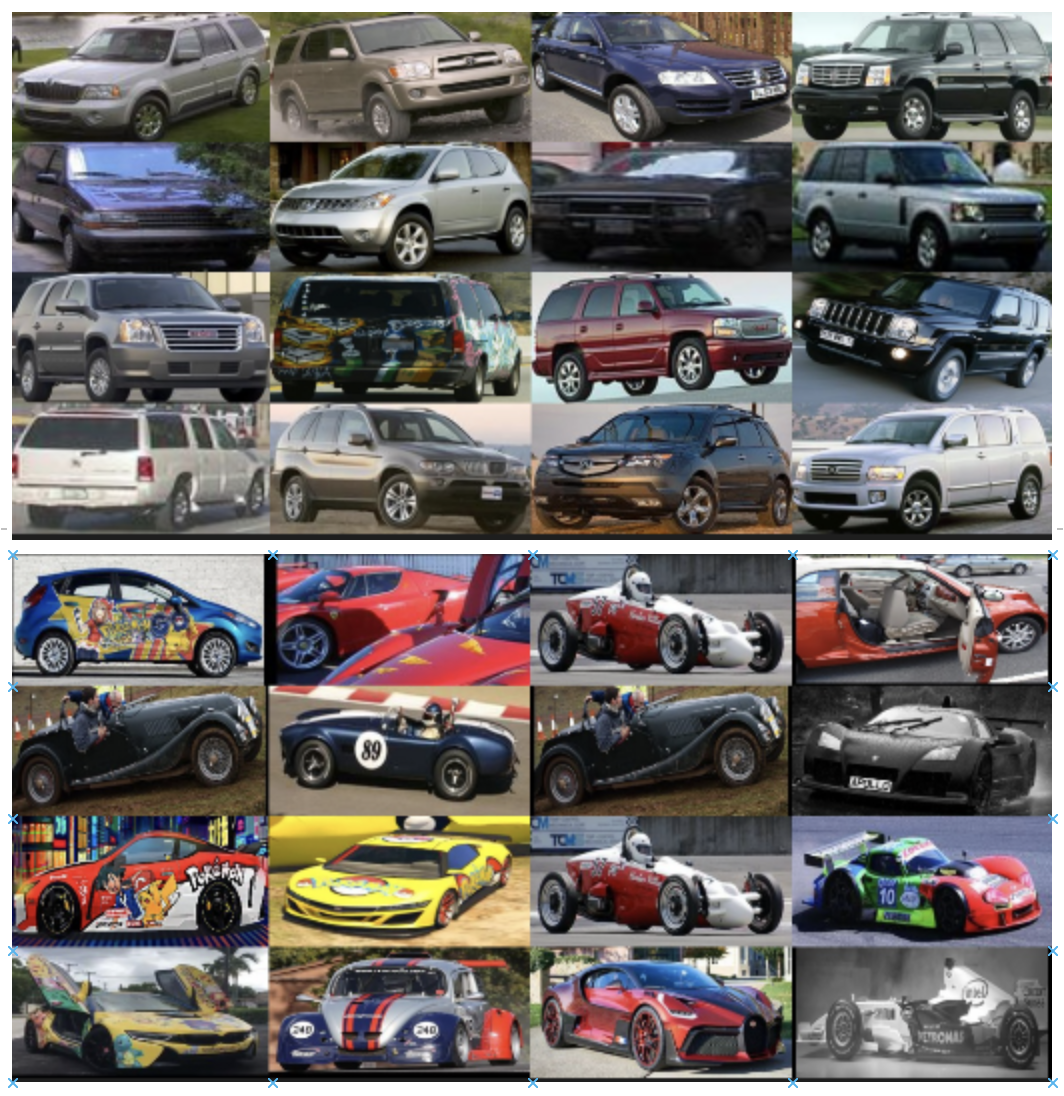}
    \caption{Subset of images from OOD-CV dataset activated by the same transitional spatial coefficient for clean source (top) data and nuisance ridden target (bottom) data.}
    \label{fig:robinmixeg}
\end{figure*}

\subsubsection{Training Details}\label{train_details}
For RPL\cite{rpl} and BNA\cite{bna}, we used the official implementation\cite{robusta}. For CompNets\cite{compnet2020journal} as well, we used the official implementation\cite{comps}. For CompNets\cite{compnet2020journal} and our model, to get feature vectors, we used a model backbone trained on the OOD-CV\cite{zhao2021robin} training data. We use VGG16 (with Batch Normalization) and Residual Network (resnet50) for all models as model backbones as most aforementioned works use them. For CompNets\cite{compnet2020journal} and our model, we use $\kappa=30$ (vMF variance) and the length of vMF dictionary is 512, with each vMF dictionary vector being 1x512. The initial VGG16 and Resnet50 backbones are trained using OOD-CV training data with 90-10 training-validation data split, and the same backbones are used for all the methods. These baseline backbones are trained using stochastic gradient descent optimizer with an initial learning rate of .0001 and weight decay of .0005. The batch size is 64 and the learning rate is changed by an exponential learning rate scheduler which decays the learning rate of each parameter group by gamma$=0.95$ every 15 epochs. We use the same optimizer and learning rate schedules for our BNA\cite{bna} and RPL\cite{rpl} experiments. For the generalized cross entropy loss\cite{gce}, we use the recommended hyperparameter value $q=0.8$ in all experimental usage. Hyperparameters $\psi_v = \psi_{\alpha} =3$ for both CompNets and \textit{UGT}. For finetuning our model, we freeze the model backbone parameters.

For MCC~\cite{mcc19}, CDAN~\cite{cdan18}, MCD~\cite{mmcd18}, MDD~\cite{mdd19} and BSP~\cite{bsp19}, we use the Transfer Learning library\cite{transferlibrary} implementations. We use the recommended hyperparameters for each method. Although, we do not use an Imagenet pretrained backbone for CompNets\cite{compnet2020journal}, RPL\cite{rpl}, BNA\cite{bna} and our method (UGT) as an Imagenet trained backbone would provide an unfair advantage to the method in the task of OOD robustness for an unlabelled data. However, for MCC~\cite{mcc19}, CDAN~\cite{cdan18}, MCD~\cite{mmcd18}, MDD~\cite{mdd19} and BSP~\cite{bsp19}, which are all well known Unsupervised Domain Adaptation methods, the model performance is not good when training from scratch. And therefore, we relax this pretrained backbone constraint for these methods, and show that even though our method is not using an Imagenet backbone, we are still able to perform better than current state-of-the-art OOD robustness\cite{rpl, bna} and unsupervised domain adaptation methods~\cite{mcc19, cdan18, mmcd18, mdd19, bsp19}.
\begin{table*}
\begin{center}
\caption{Hyperparameters}
\label{table:hyper}
\begin{tabular}{llccccc}
\hline\noalign{\smallskip}
\textbf{Model}       & \textbf{Backbone} & epochs & lr & Batch  & \textbf{layer}  \\
\noalign{\smallskip}
\hline
\noalign{\smallskip}
\textbf{Compnet\cite{compnet2020journal}} & \textbf{VGG16}    & 60 & 0.01 & 64 & pool5   \\
\textbf{RPL\cite{rpl}}     & \textbf{VGG16}    & 60 & 0.0001  & 64 &  -  \\
\textbf{BNA\cite{bna}}     & \textbf{VGG16}    & 60 & 0.0001 & 64 &  - \\
\textbf{UGT (Ours)}    & \textbf{VGG16}    &  50  & 0.01  & 64 &  pool5  \\
\hline
\textbf{Compnet\cite{compnet2020journal}} & \textbf{Resnet50} & 60 & 0.01 & 64  & layer4   \\
\textbf{RPL\cite{rpl}}     & \textbf{Resnet50} & 60 & 0.0001 & 64 &  - \\
\textbf{BNA\cite{bna}}     & \textbf{Resnet50} & 60 & 0.0001 & 64 &  -  \\
\textbf{UGT (Ours)}    & \textbf{Resnet50} & 50 & 0.01 & 64  & layer4 \\
\hline
\end{tabular}
\end{center}
\end{table*}
\subsection{Imagenet-C}
\begin{figure*}[!h]
    \centering
    \includegraphics[height=4cm]{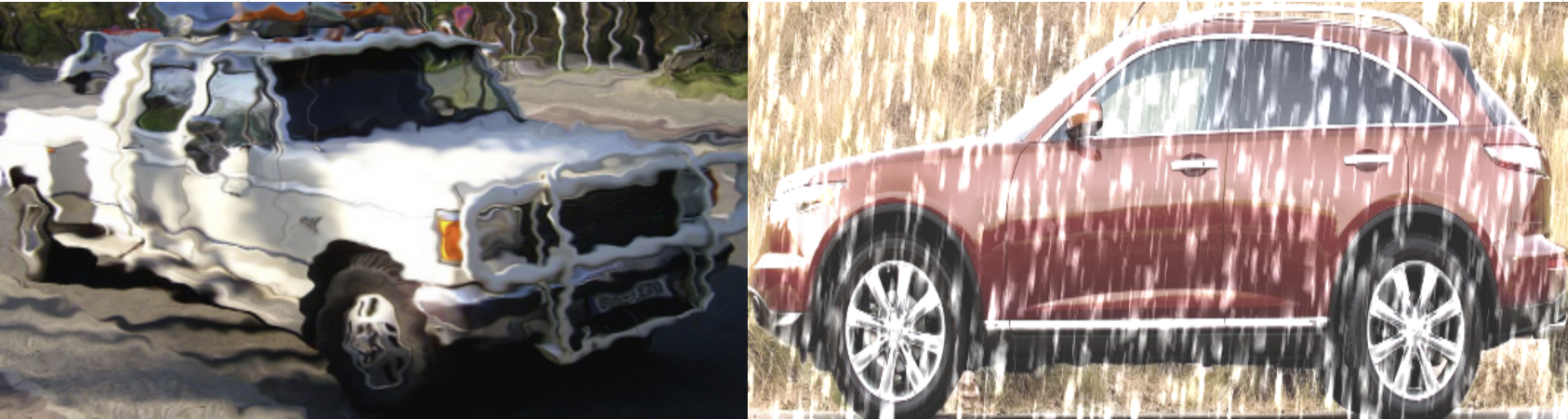}
    \caption{Example Images with Snow and Elastic corruptions (Severity 4) from Imagenet-C}
    \label{fig:imc}
\end{figure*}
Since previous OOD robustness methods~\cite{rpl, bna} primarily use synthetic corruptions~\cite{imagenetC}, we employ the same corruptions and apply them to objects from Pascal3D+ dataset. The objective is to robustly adapt to an unlabelled data of synthetically corrupted images when the training data consists of only clean images. We apply synthetic corruptions to Pascal3D+ test data for our evaluation. Tables~\ref{table:snow}, \ref{table:gaussnoise}, \ref{table:shotnoise}, \ref{table:pixelate} show additional classification accuracy results on different corruption target dataset versions.

\snowu
\gaussnoiseu
\shotnoiseu
\pixelateu
\glassbluru
\textbf{Training Details} We follow the same training methodology as mentioned in Section~\ref{train_details}.

\subsection{UDAParts (Synthetic to Real)}
\begin{figure*}[!htb]
    \centering
    \includegraphics[height=8cm]{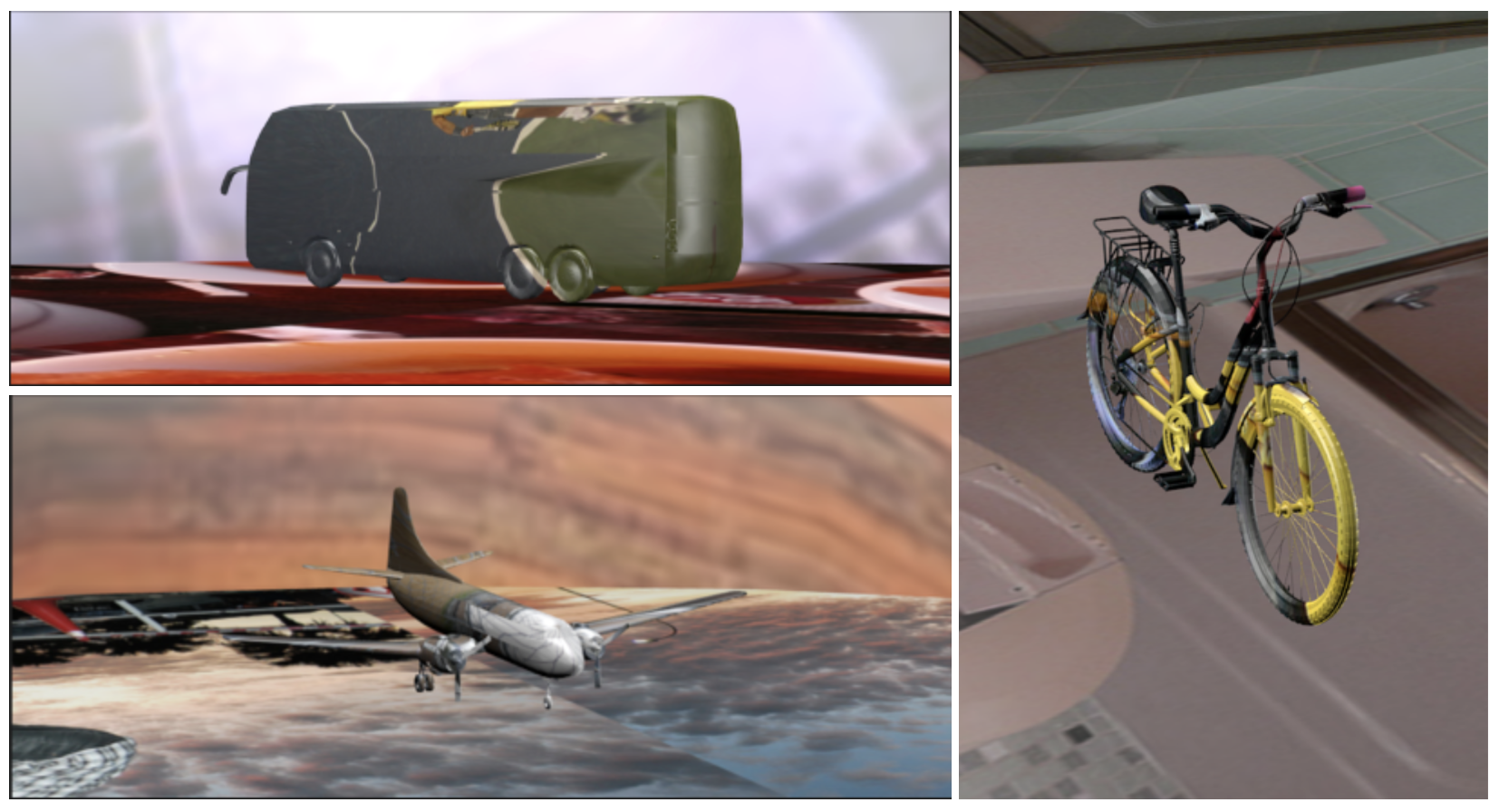}
    \caption{Examples from UDAParts dataset}
    \label{fig:udaparts2}
\end{figure*}

Figure~\ref{fig:udaparts2} shows an example of the synthetically created images that are used as our training data. They have created using object models and rendered using random 3D pose, texture and background. The purpose of this task is to make the model (which has trained on the synthetic data) robust to the OOD real data, which in this case is Pascal3D+. Additionally, the model needs to be robust to OOD real data which suffers from partial occlusion\cite{occludedpascal}. We choose around 2000 random images from the randomly generated images in UDAParts for each class of object as our training set.
\begin{figure*}[!htb]
    \centering
    \includegraphics[height=10cm]{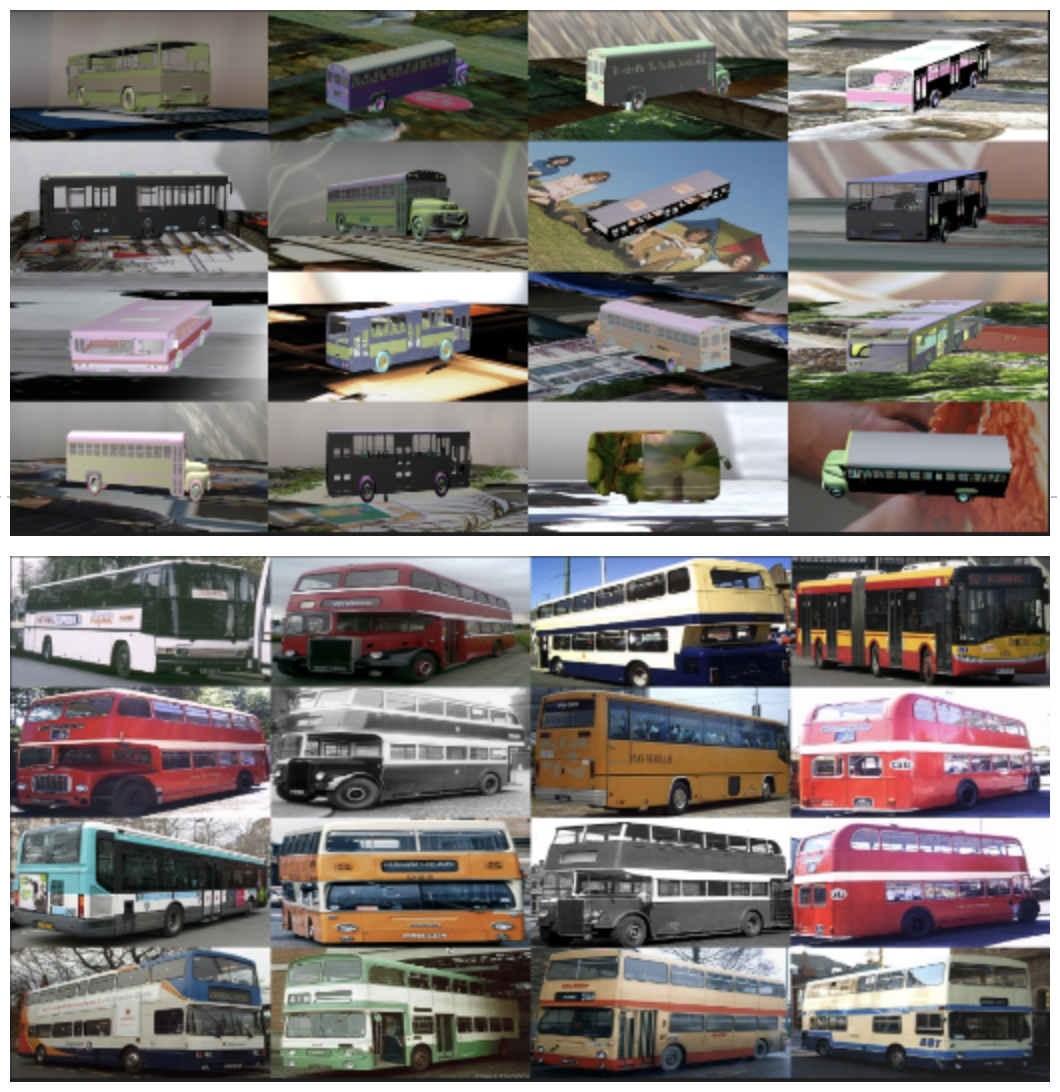}
    \caption{Subset of images activated by the same transitional spatial coefficient for synthetic source (top) data and real target (bottom) data.}
    \label{fig:udamix}
\end{figure*}
\textbf{Training Details} We follow the same training methodology as mentioned in Section~\ref{train_details}.\\
Figure~\ref{fig:udamix} shows an example of images which are activated by the same transitional spatial coefficient in the source (UDAParts) and target (Pascal3D+) dataset.

\subsection{Data Augmentation and Model Backbones}
We avoid using different data augmentations to ensure a fair and balanced comparison between all methods. Though, additional data or different kinds of data augmentations do help alleviate the problem, we focus on the methodology aspect and avoid using various kinds of data augmentations or additional data for all methods. Similarly, we keep the model backbones' similar, whenever possible to allow for a fairer comparison.

\subsection{Training Cost}
Training Costs (time) is similar to the baseline generative model~\cite{compnet2020journal} as we are only finetuning a few layers(spatial mixture matrix). Our code isn’t optimized yet and our method, sans the finetuning, uses only the CPU. On AMD 5700x CPU and for a training data size of 10k samples,it takes  $~6$ hrs to learn the transitional model and 15 minutes to finetune the model for 10 epochs on a Nvidia 3070.

\section{Ablation}
As we show in our ablation results in the main draft, each individual improvement of our method, UGT, markedly improves the classification results for unsupervised domain robustness. Here, 
$\Lambda^{\mathcal{R}}+\mathcal{A^R}$ refers to calculating transitional vMF dictionary and then calculating the spatial coefficients using this transitional dictionary and source feature vectors. $\mathcal{A}^{\mathcal{R'}}$ refers to the finetuned transitional spatial coefficient.


In Tables~\ref{table:ablationgaussnoise},~\ref{table:ablationpixelate},~\ref{table:ablationmotion},~\ref{table:ablationshot} and ~\ref{table:ablationgaussblur}, we can see the improvements in Imagenet-C experiments where different kinds of Imagenet-C corruptions are applied to objects from Pascal3D+ dataset.
\ablationgaussnoise
\ablationpixelate
\ablationmotion
\ablationshot
\ablationgaussblur
\ablationelastic

 We can notice that just learning the transitional parameters is often enough to drastically improve model's performance underlying its importance in our work.
\section{Transitional vMF dictionary}
The transitional vMF dictionary is learnt by initializing the dictionary vectors with learnt $\Lambda^{\mathcal{S}}$ and regularizing the learning of $\Lambda^{\mathcal{R}}$ by the cosine distance between $\mu^\mathcal{S}$ and $\mu^\mathcal{R}$. Since, the directional parameter or the vMF variance ($\sigma_k/\kappa$) is constant (and therefore, the normalization term in the denominator is constant), $\mu$ is the only parameter being learnt. Figure~\ref{fig:three graphs} gives an intuition regarding the source and transitional vMF kernels. 

\begin{figure}[!htb]
    \centering
    \includegraphics[height=4.5cm]{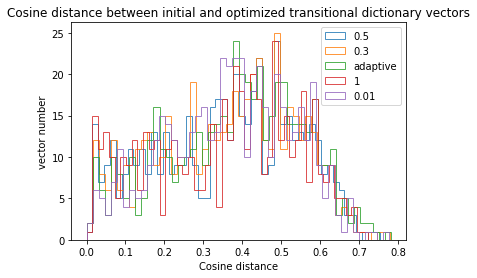}
    \caption{Histogram representing the cosine distance between initial and final transitional dictionary vectors with different values of $\psi_k$}
    \label{fig:occcos}
\end{figure}

Also, as we can see in Figure~\ref{fig:occcos}, the choice of the adaptation coefficients $\psi_k$, which represents the hyperparameters used to tune the adaptation between source and transitional $\mu_k$, is not that critical. Figure~\ref{fig:occcos} represents the learning of vMF kernel dictionary for the combined nuisance experiment where the entire OOD-CV test data is considered the target domain. 
All the normalized $\psi_k$ are initialized to an arbitrary value between 0 and 1 and then step wise increased until likelihood improvement of the particular vMF dictionary vector is below a threshold value($1e-5$). 
As seen in the figure, different initial values for the adaptation coefficient produce similar results and therefore the usage of the data dependent/adaptive coefficient is optional in our experiments.

\begin{figure}[!htbp]
    \centering
    \includegraphics[height=14cm]{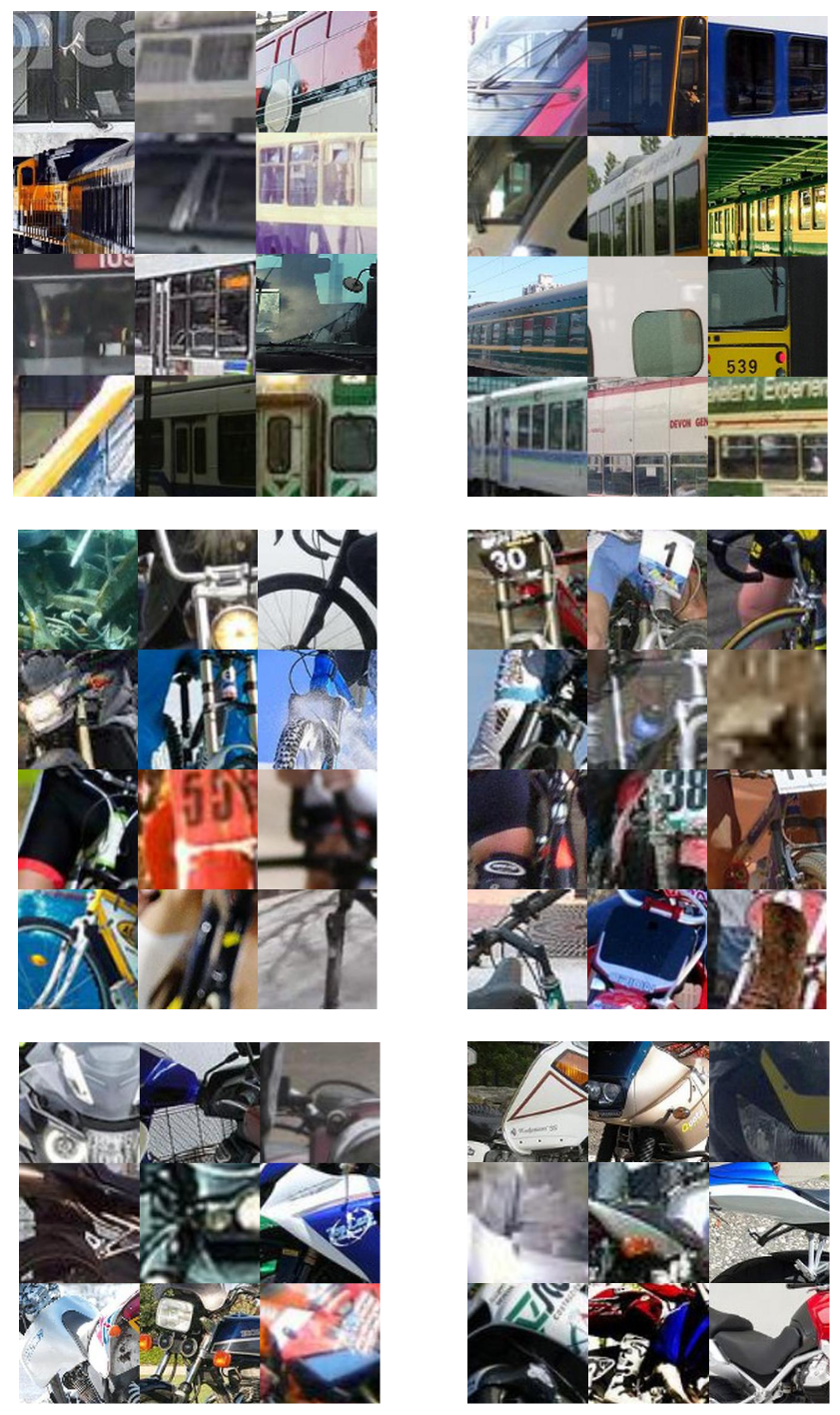}
    \caption{Image patches from OOD-CV training (left) and nuisance test (right) data which have high activations for corresponding vMF vectors from the source (initialised $\Lambda^{\mathcal{R}}$) and transitional vMF dictionaries.}
    \label{fig:t1}
\end{figure}

\begin{figure*}[!htb]
    \centering
    \includegraphics[height=9cm]{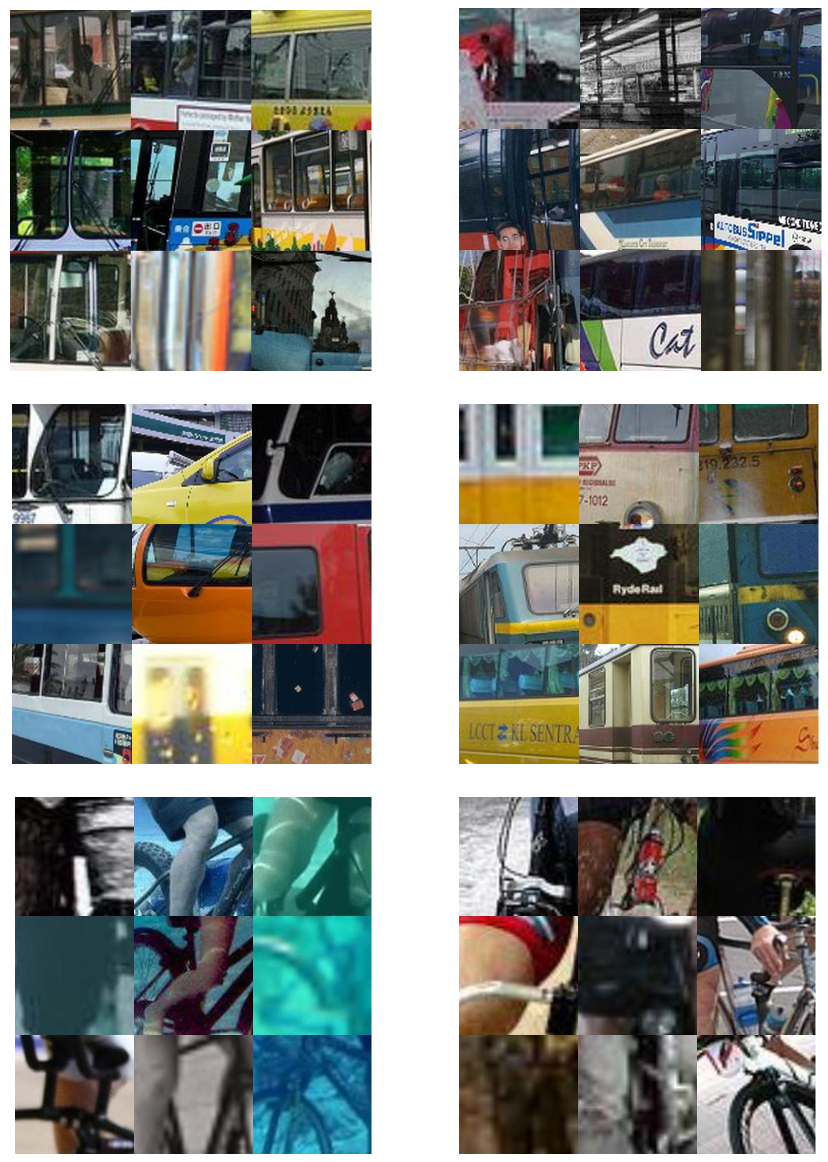}
    \includegraphics[height=9cm]{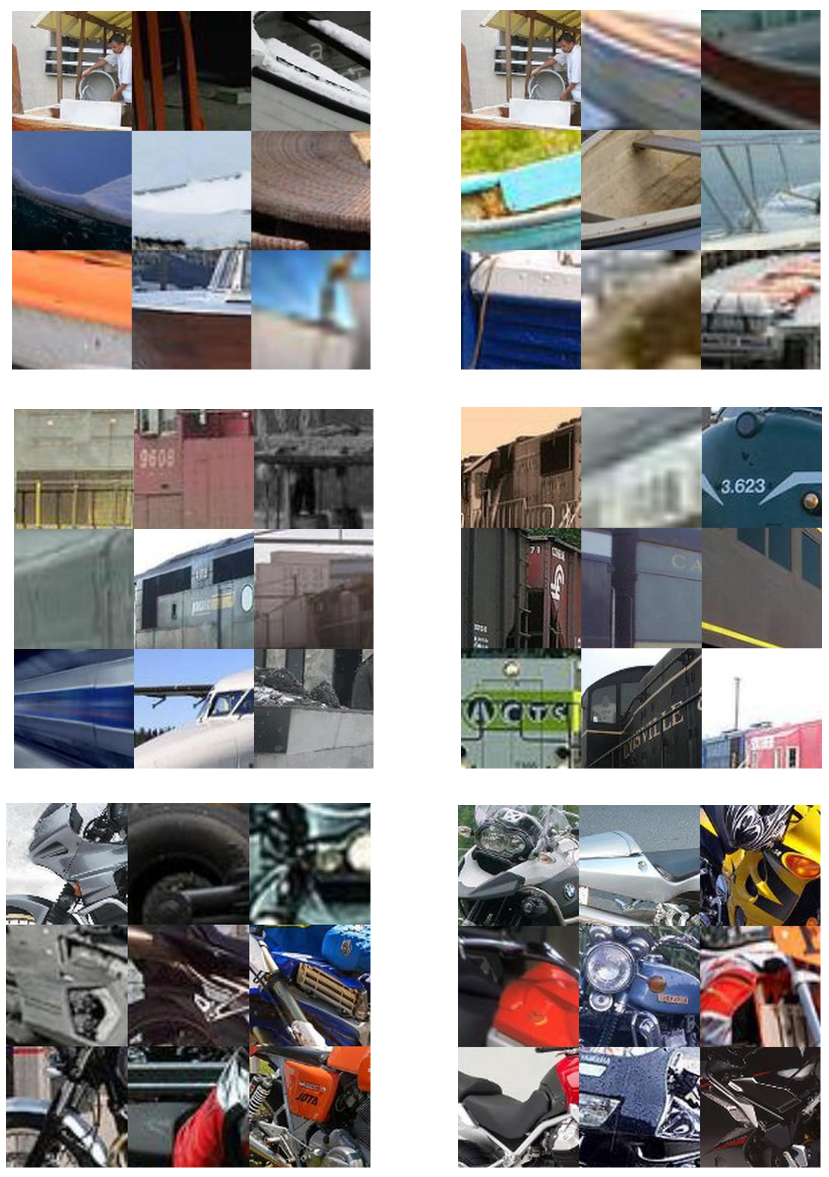}
    \caption{Additional examples of roughly corresponding image patches from OOD-CV training (left) and nuisance test (right).}
    \label{fig:t2-3}
\end{figure*}

\begin{figure*}[!htb]
    \centering
    \includegraphics[height=6cm]{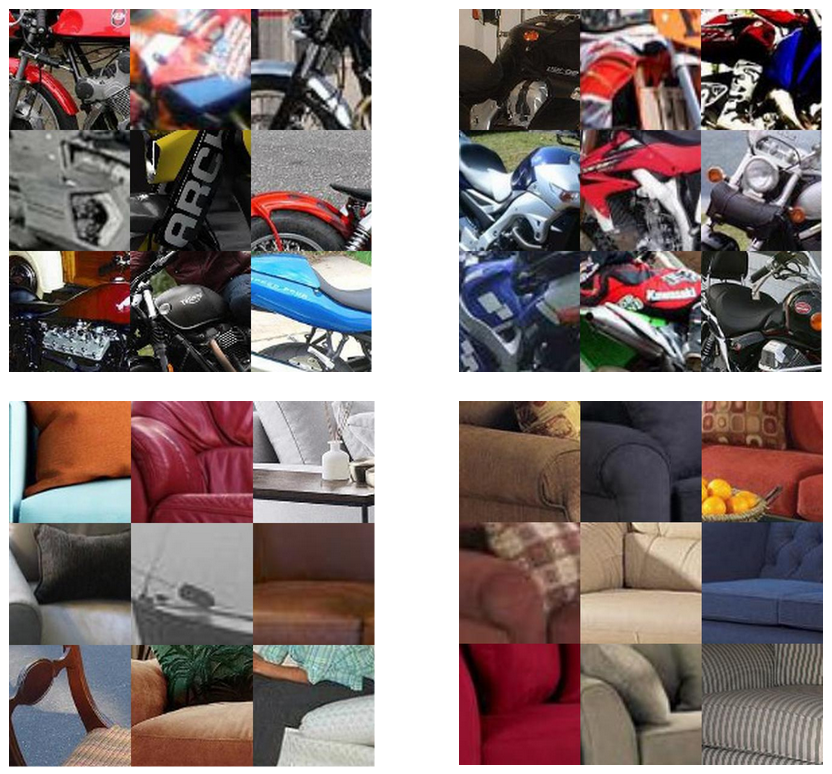}
    \caption{Additional example image patches from OOD-CV training (left) and nuisance test (right) data.}
    \label{fig:t4}
\end{figure*}

Figures~\ref{fig:t1},\ref{fig:t2-3} and \ref{fig:t4} show examples of source and their corresponding transitional vMF dictionary vector image patches. The image patches roughly correspond to parts of an image in the data which has high activations for a specific vMF dictionary vector. The image patches are from the training and evaluation sets of OOD-CV dataset respectively, which have been curated to filter out training like data in the nuisance factored test set. However, we can still notice that the approximate image patches actually correspond to similar object parts despite drastic changes in 3D pose, shape, texture, weather, etc. For e.g, in Figure~\ref{fig:t1}, we can see patches focusing on vehicle windows, bicycle and motorbike handlebars, etc. 
\subsection{Finetuning via Pseudo-Labelling}
We finetune our transitional spatial coefficients by simple pseudo-labelling on a target data subset. This subset is not used during evaluation - although test time finetuning can be performed to improve the model's performance during inference. We do not follow any sophisticated pseudo-labelling methods and simply use a high threshold value (e.g 0.8) for the final class activations to pseudo-label target domain samples. We can, presumably, achieve higher performance using better and more sophisticated pseudo-labelling and finetuning methods, however, this is not a point of concern for this work.


\section{Future Work and Limitations}
\begin{figure*}[!htbp]
  \centering
  \includegraphics[width=15cm]{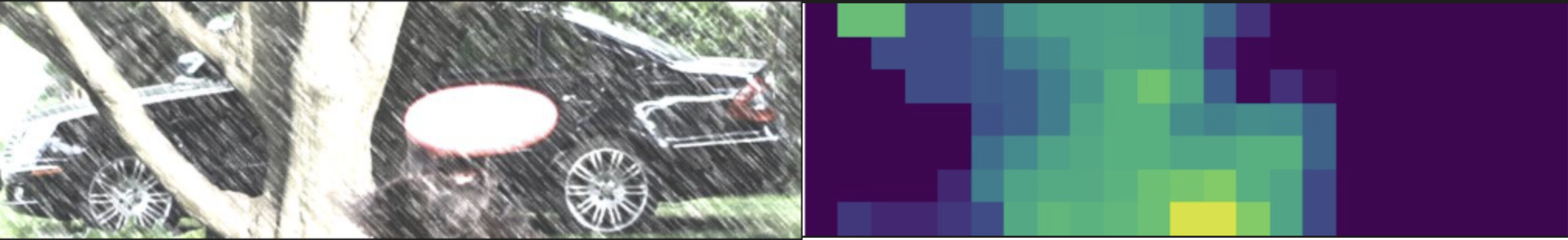}
   \caption{Occluder localization in a \textit{snow} corrupted car image}
   \label{fig:heatmap}
\end{figure*}
Figure~\ref{fig:heatmap} shows that our adapted image classification model is capable of localizing occluders (like CompNets\cite{compnet2020journal} do in a non-corrupted scenario) even in a nuisance ridden target image. Thus, we think that applications like object detection and segmentation are interesting future applications of our work.

In terms of limitations, our method may not work in case of extreme divergence between the source and target domains which may be be caused by inter-domain noise or large difference in both domain object parts and their spatial orientations. Our model is also bound to the assumptions and weaknesses of the Bayesian, generative model~\cite{Kortylewski2020CompositionalCN} which we build upon, a discussion of which can be found in the relevant works, like~\cite{compnet2020journal, kortylewski2021compositional, sun2020weakly}.
 











\clearpage
{\small
\bibliographystyle{ieeenat_fullname}
    \bibliography{egbib}
}